\documentclass{article}

\usepackage{microtype}
\usepackage{graphicx}
\usepackage{booktabs} % for professional tables

\usepackage{hyperref}

\usepackage[accepted]{conferences}

\usepackage{amsmath}
\usepackage{amssymb}
\usepackage{mathtools}
\usepackage{amsthm}

\usepackage{multicol}
\usepackage{wrapfig}

\usepackage[capitalize,noabbrev]{cleveref}

\theoremstyle{plain}

\theoremstyle{definition}

\theoremstyle{remark}

\usepackage[textsize=tiny]{todonotes}

\usepackage{multirow}
\usepackage{bm}
\usepackage{bbm}
\usepackage{wrapfig}
\usepackage{xcolor}
\usepackage{color, colortbl}
\usepackage{tabu}
\usepackage{subcaption}

\newcommand{\cmark}{\checkmark}
\newcommand{\xmark}{$\times$}

\definecolor{gg}{gray}{0.70}

\definecolor{baselinecolor}{gray}{.9}

\icmltitlerunning{E-MD3C: Taming Masked Diffusion Transformers for Efficient Zero-Shot Object Customization}

\begin{document}

\twocolumn[
\icmltitle{E-MD3C: Taming Masked Diffusion Transformers for Efficient Zero-Shot Object Customization}




\begin{icmlauthorlist}
% \icmlauthor{Trung X. Pham}{equal,yyy}
% \icmlauthor{Zhang Kang}{equal,yyy}
% \icmlauthor{Chang D. Yoo}{yyy}\\
\icmlauthor{Trung X. Pham}{}
\icmlauthor{Zhang Kang}{}
\icmlauthor{Ji Woo Hong}{}
\icmlauthor{Xuran Zheng}{}
\icmlauthor{Chang D. Yoo}{}\\
KAIST\\
School of Electrical Engineering
\end{icmlauthorlist}
% \icmlaffiliation{comp}{Korea Advanced Institute of Science and Technology (KAIST)}
% \icmlaffiliation{yyy}{School of Electrical Engineering, KAIST, South Korea.}

\icmlcorrespondingauthor{Trung X. Pham}{trungpx@kaist.ac.kr}
\icmlcorrespondingauthor{Chang D. Yoo}{cd\_yoo@kaist.ac.kr}



\icmlkeywords{Machine Learning, Computer Vision, Object Customization, Highly Efficient, Masked Diffusion Models}

\vskip 0.3in
]



\printAffiliationsAndNotice{\icmlEqualContribution} % otherwise use the standard text.

% This document provides a basic paper template and submission guidelines.
% Abstracts must be a single paragraph, ideally between 4--6 sentences long.
% Gross violations will trigger corrections at the camera-ready phase.
\begin{abstract}
We propose \texttt{E-MD3C} (\underline{E}fficient \underline{M}asked \underline{D}iffusion Transformer with Disentangled \underline{C}onditions and \underline{C}ompact \underline{C}ollector), a highly efficient framework for zero-shot object image customization. Unlike prior works reliant on resource-intensive Unet architectures, our approach employs lightweight masked diffusion transformers operating on latent patches, offering significantly improved computational efficiency. The framework integrates three core components: (1) an efficient masked diffusion transformer for processing autoencoder latents, (2) a disentangled condition design that ensures compactness while preserving background alignment and fine details, and (3) a learnable Conditions Collector that consolidates multiple inputs into a compact representation for efficient denoising and learning. \texttt{E-MD3C} outperforms the existing approach on the VITON-HD dataset across metrics such as PSNR, FID, SSIM, and LPIPS, demonstrating clear advantages in parameters, memory efficiency, and inference speed. With only $\frac{1}{4}$ of the parameters, our Transformer-based 468M model delivers $2.5\times$ faster inference and uses $\frac{2}{3}$ of the GPU memory compared to an 1720M Unet-based latent diffusion model.
\end{abstract}

\vspace{-18pt}
\section{Introduction}
\label{intro_sec}
% \begin{enumerate}
%     % \item New VAE DCAE $\rightarrow$ ICLR 2025 paper (\textcolor{red}{DONE, slow convergence, move back to SDAE})
%     % \item Multiple level features DINOv2 all with learnable MLP alignment $\rightarrow$ ICLR 2025 paper (REPA)
%     % \item FLOW MATCHING applying to the first method on this task (\textcolor{red}{CONSIDER LATER, SiT})
%     % \item Minimize the attention to the interested region of the object
%     % \item Crop region of the salient object to get the best details
%     \item MDT $\Rightarrow$ Zero-shot task (\textcolor{blue}{DONE})
%     \item Background hinting (\textcolor{blue}{DONE})
%     \item Conditions collector (\textcolor{blue}{DONE})
%     \item Cross-attention at the beginning of the VAE visual tokens (\textcolor{blue}{DONE})
%     \item Adding DINO hint for more accurate prediction (\textcolor{blue}{DONE})
%     \item MDT models with masking more on object regions (\textcolor{blue}{DONE}) checking mask ratio.
% \end{enumerate}

\begin{figure}%[!ht] %[!htbp]
  \centering
  \includegraphics[width=1\linewidth]{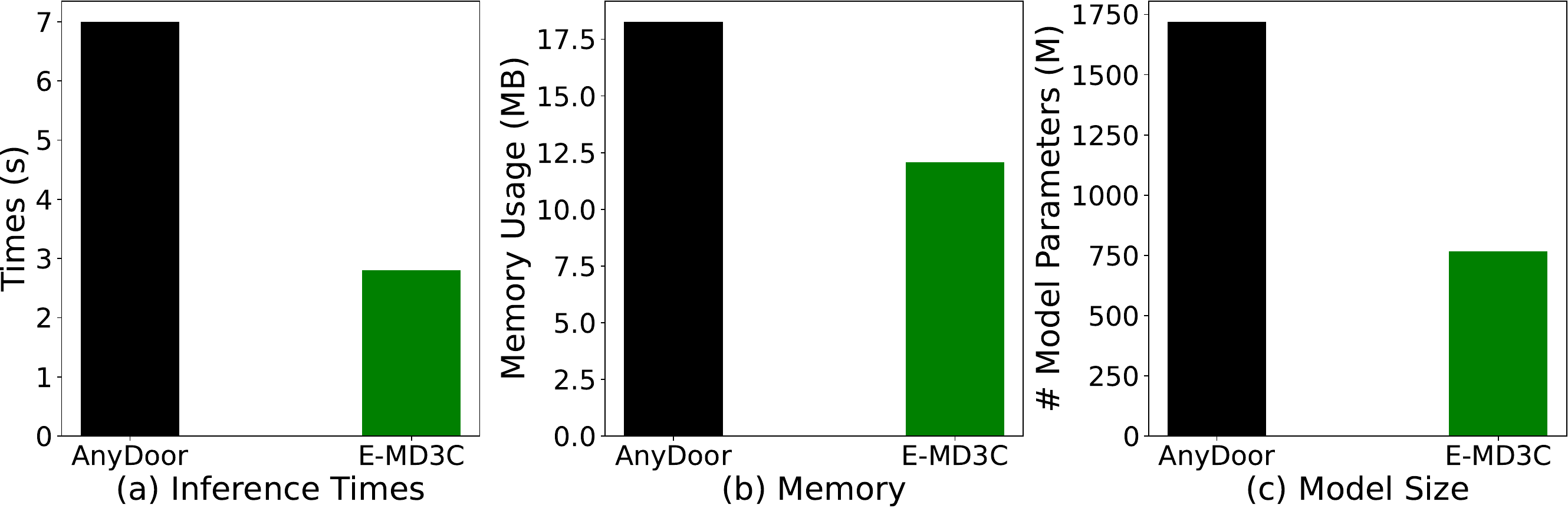}
  \vspace{-14pt}
  \caption{\textbf{Existing Approach Inefficiency.} The current model (black) demands significant parameters, memory, and inference time due to its resource-intensive architecture.}
  \label{fig:efficiency_bar}
  \vspace{-18pt}
\end{figure}

We address the challenge of zero-shot object-level image customization (ZSOIC) \cite{chen2024anydoor}, a task that has gained traction with the rise of diffusion models and open-source frameworks like Stable Diffusion \cite{rombach2022high, zhang2023adding}. Existing methods, including ObjectStitch \cite{song2023objectstitch} and Paint-by-Example \cite{yang2023paint}, focus on localized edits but struggle with identity consistency, especially for unseen categories. Customization techniques such as Textual Inversion \cite{gal2022image}, DreamBooth \cite{ruiz2023dreambooth}, and UniCanvas \cite{jin2025unicanvas} enable novel concept generation but lack spatial control and require extensive fine-tuning, limiting their real-world applicability. AnyDoor \cite{chen2024anydoor} mitigates these issues by introducing a Stable Diffusion-based pipeline that ensures ID consistency in zero-shot settings, achieving flexible, high-quality region-specific edits using ID tokens and frequency-aware detail maps. However, its reliance on resource-heavy Unet-based architectures with ControlNet \cite{zhang2023adding} significantly increases computational overhead (see Fig. \ref{fig:efficiency_bar}). Notably, prior works have largely overlooked the efficiency and potential of Vision Transformers (ViTs) \cite{dosovitskiy2021an}, which offer a compelling alternative to Unet-based designs. This raises a key question: \textit{Can we achieve competitive performance while substantially reducing computational costs?}

Inspired by recent advances in transformer-based diffusion models, such as DiT \cite{peebles2023scalable, esser2024scaling, xie2025sana} and masked diffusion transformers \cite{gao2023masked, phamcross, mao2024mdt, pham2025mdsgen}, we explore their efficiency and adaptability for ZSOIC. Diffusion transformers \cite{peebles2023scalable} offer stable training and strong generative performance, outperforming traditional Unet-based approaches \cite{ho2020denoising, rombach2022high}, making them well suited for this task. We introduce \texttt{E-MD3C}, a novel framework that employs masked diffusion transformers \cite{gao2023masked} for ZSOIC. Our approach features a specialized disentangled collection network that consolidates conditions into a compact vector for adaptive LayerNorm modulation, alongside a masking network to enhance transformer learning. Despite its simple yet effective design, E-MD3C surpasses existing methods in FID, SSIM, L1, and LPIPS on the VITON-HD dataset \cite{choi2021viton}, while significantly reducing parameter count and computational demands. Our main contributions are as follows:
\begin{itemize}
    \vspace{-6pt}
    \item We propose \texttt{E-MD3C}, the first masked diffusion transformer-based model for zero-shot object customization, achieving high efficiency through latent patches and compact multi-condition representations.
    \vspace{-6pt}
    \item A novel Conditions Collector Module (CCM) is introduced to consolidate essential information into a lightweight representation for denoising diffusion, significantly reducing computational complexity compared to methods that concatenate full conditions with noisy target images.
    \vspace{-6pt}
    \item Our framework decouples conditions into two branches: the hint image supervises the target image in the denoising branch, while other conditions are processed via the CCM and a single cross-attention layer. This design enhances token alignment, preserves background details, and ensures robust correspondence between target and source images.
    \vspace{-6pt}
    \item \texttt{E-MD3C} outperforms the heavy Unet-based approach across multiple metrics, using only a quarter of the parameters and achieving $2.5\times$ faster inference.
\end{itemize}

\section{Related Works}
\label{related_sec}
\paragraph{Zero-shot Object Customization.}
\citet{chen2024anydoor} introduced AnyDoor to tackle the ZSOIC problem, utilizing a large pre-trained diffusion model, Stable Diffusion \cite{rombach2022high}, with a dual-backbone architecture inspired by ControlNet \cite{zhang2023adding}. Despite achieving strong performance, AnyDoor relies on a resource-intensive architecture with significant computational demands, making the image generation process inefficient. Similarly, \citet{chen2024zeroshot} proposed MimicBrush, a Unet-based latent diffusion model that integrates a reference U-Net into an imitative U-Net, offering finer part-level control but requiring a large model size and heavy computational resources. UniCanvas \cite{jin2025unicanvas} also employs a substantial framework built upon customized text-to-image generation with Stable Diffusion, introducing a novel affordance-aware editing pipeline but further amplifying computational overhead with its parallel generative branches. While these approaches improve object customization applications, their reliance on oversized CNN-based architectures limits practical usage and applicability, especially on hardware with limited resources. To address these shortcomings, we propose \texttt{E-MD3C}, a lightweight transformer-based diffusion model that generates high-quality images while significantly reducing resource requirements (see Fig. \ref{fig:efficiency_bar}).

\begin{figure}[!htbp]
  \centering
  \includegraphics[width=1.0\linewidth]{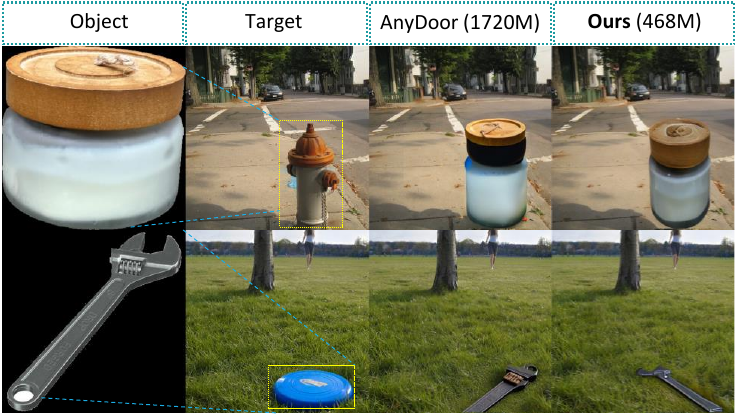}
  \vspace{-10pt}
  \caption{\textbf{Object Composition}. The 3$^{\text{rd}}$ and 4$^{\text{th}}$ columns show outputs from the existing method and our model. Our model generates images in just over 2 seconds, compared to 7 seconds for the existing approach.}
  \label{fig:DreamBooth}
\vspace{-14pt}
\end{figure}

\paragraph{Denoising Diffusion Transformers.}
The CNN-based U-Net architecture \cite{ronneberger2015u} has long been the foundational framework for diffusion models and remains a prevalent choice for various diffusion-based generation tasks \cite{ho2020denoising, ho2022cascaded, song2019generative}. However, a transformative shift occurred with the introduction of DiT \cite{peebles2023scalable}, which incorporated the transformer-based ViT architecture \cite{dosovitskiy2021an} into latent diffusion models. This innovation demonstrated superior scalability and consistently outperformed traditional U-Net-based designs. Building on this progress, \citet{gao2023masked} further advanced diffusion transformers, achieving state-of-the-art results in class-conditional image generation on ImageNet through advanced contextual representation learning. Other works, such as Stable Diffusion v3 \cite{esser2024scaling}, PixArt \cite{chen2025pixart} and Sana \cite{xie2025sana}, have leveraged DiT for text-to-image generation, highlighting the strong potential of transformer architectures in generative tasks. Although prior studies primarily focus on general-purpose or text-driven generative tasks, our research investigates the application of diffusion transformers to the specialized and complex task of zero-shot object-level image customization (ZSOIC). ZSOIC involves extracting and integrating intricate attributes from a source image, including object appearance, identity, and background, into a unified target image, which poses unique challenges in achieving cohesive synthesis \cite{ginesu2012objective, chen2024anydoor} (See Fig. \ref{fig:DreamBooth}).

\begin{figure*} %[!htbp]
  \centering
  \includegraphics[width=1\linewidth]{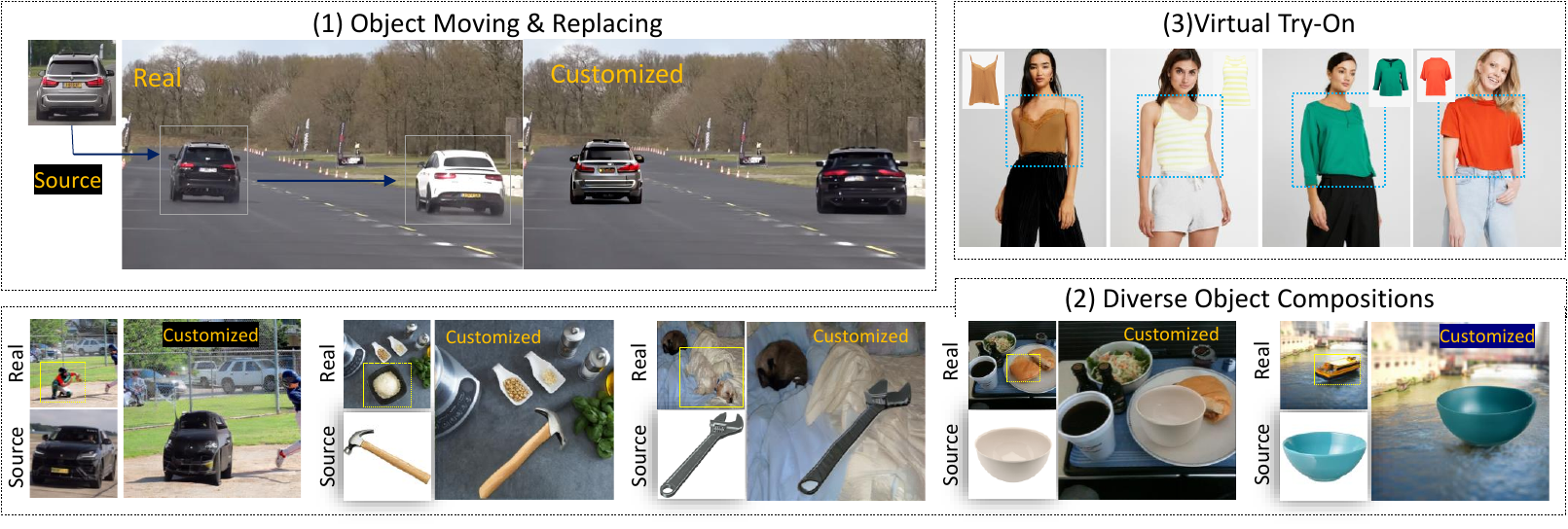}
  \vspace{-16pt}
  \caption{Zero-shot object customization and its practical applications. Images are generated using our \texttt{E-MD3C} model.}
  \label{fig:task}
  \vspace{-12pt}
\end{figure*}

\paragraph{Mask Prediction Modeling.}
Mask-based vision models, drawing inspiration from mask language models like BERT \cite{devlin2018bert}, have demonstrated exceptional scalability and performance across diverse tasks. Prominent examples include MAE \cite{he2022masked} for self-supervised learning (SSL) and models such as MaskGIT \cite{chang2022maskgit}, Muse \cite{chang2023muse}, and MAGVIT \cite{yu2023magvit}, which learn discrete token distributions for image generation. These methods adopt non-autoregressive modeling with a cross-entropy loss to predict token indices within a codebook. Diverging from this paradigm, MDT \cite{gao2023masked} introduced an asymmetric masking schedule that enhances contextual representation learning in denoising diffusion transformers, achieving superior class-conditional image generation on ImageNet. Building on this foundation, \citet{phamcross} proposed X-MDPT, demonstrating the potential of masked diffusion transformers for pose-guided human image generation with improved performance and efficiency. Similarly, MDT-A2G \citep{mao2024mdt} applied masked diffusion transformers to gesture generation, while QA-MDT \citep{li2024quality} extended this framework for music generation. Further advancements include masked diffusion transformers for audio generation, as explored by MDSGen \cite{pham2025mdsgen}, achieving impressive results and inspiring subsequent research. In this work, we focus on leveraging the strengths of masked diffusion transformers \cite{gao2023masked} for the task of zero-shot object-level image customization, introducing \texttt{E-MD3C}, a disentangled design with multiple conditioning mechanisms to enable effective and targeted object generation with various applications as shown in Fig. \ref{fig:task}. 

\begin{figure*}%[!htbp]
  \centering
  \includegraphics[width=0.95\linewidth]{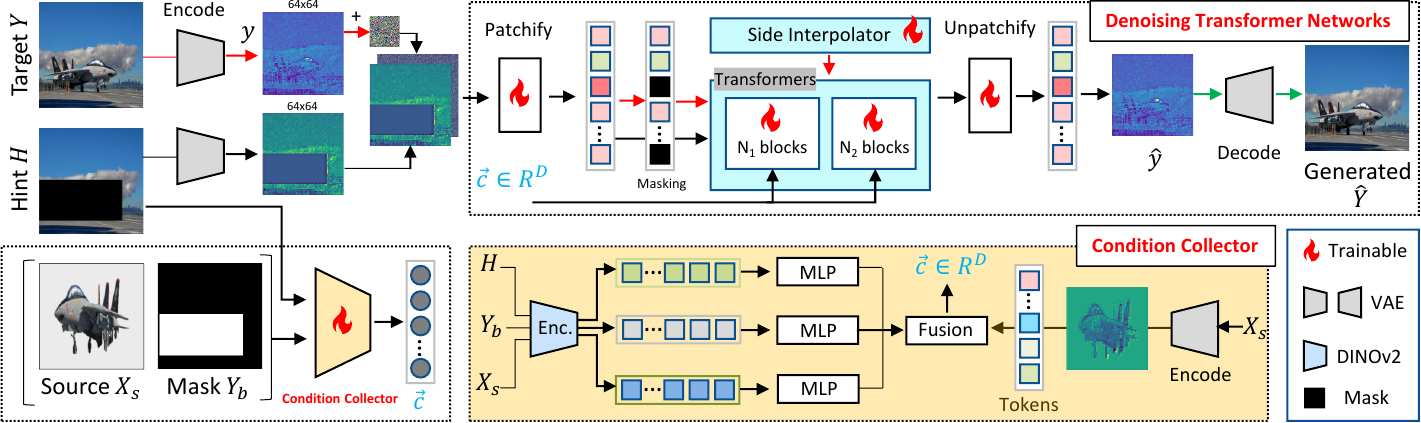}
  \vspace{-8pt}
  \caption{Overview of the \texttt{E-MD3C} framework for zero-shot object customization. During training, 30\% of patched tokens are masked, and the noisy input is processed by the Diffusion Transformer, conditioned on a collected vector ($D=1024$) via AdaLN modulation \cite{peebles2023scalable}. A mask prediction objective models token relationships. The \textcolor{red}{red arrow} $\color{red}\rightarrow$ is training-only, the \textcolor{black}{black arrow} $\color{black}\rightarrow$ is used for both training and inference, and the \textcolor{green}{green arrow} $\color{green}\rightarrow$ is inference-only.}
  \label{fig:arch}
  \vspace{-12pt}
\end{figure*}

\section{Method}
\label{method_sec}
We propose a simple yet effective framework to tackle the ZSOIC task using diffusion transformers \cite{peebles2023scalable}. The overall architecture is illustrated in Fig. \ref{fig:arch}. Our method, \texttt{E-MD3C}, is composed of three core modules: (1) \textbf{Denoising Transformer-based Diffusion Network} (DTDNet): This serves as the backbone for generating high-quality target images through iterative denoising. (2) \textbf{Disentangled Masked Diffusion Module} (DMDNet): A novel masking-based design that models contextual relationships within the predicted target image by leveraging hints and noisy latents. This masking branch acts as a powerful regularization mechanism for the transformer, enhancing its contextual learning ability. (3) \textbf{Learnable Conditions Collector} (CCNet): This module aggregates residual information from the source image and bounding box details into a compact vector, facilitating efficient denoising and improved learning performance. In the following sections, we delve into each component in detail.

\begin{figure}[!htbp]
  \centering
  \includegraphics[width=0.9\linewidth]{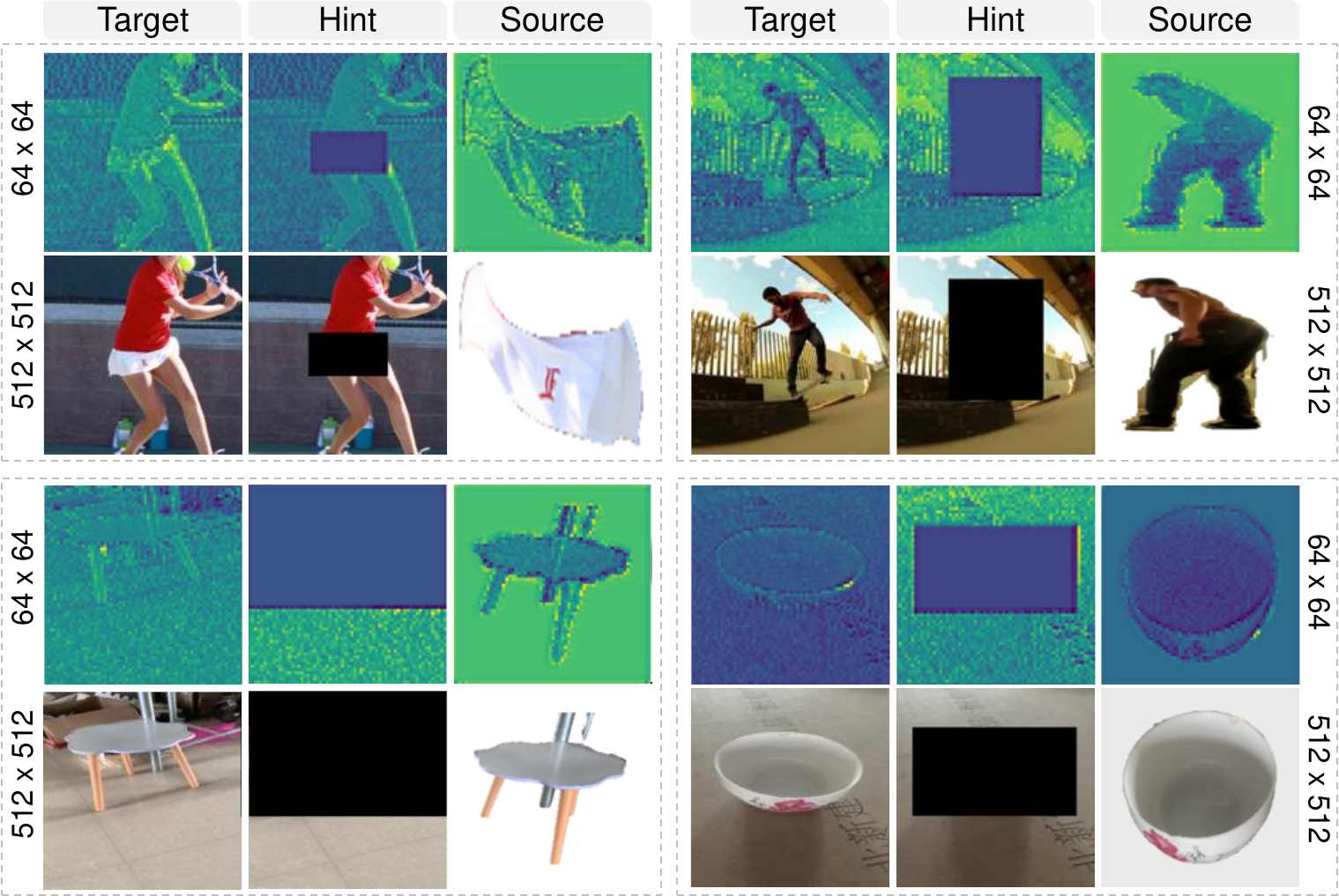}
  \vspace{-8pt}
  \caption{\textbf{Training data with diverse object sizes.} \texttt{In the pixel space} ($512 \times 512$), objects of varying sizes and positions train the model, with black areas marking masked objects in bounding boxes. \texttt{In the latent space} ($64 \times 64$), box position is preserved.}
  \label{fig:spatial_preserved}
  \vspace{-12pt}
\end{figure}

\subsection{Denoising Transformer-based Diffusion Network}
In our \texttt{E-MD3C} framework, this network component is referred to as DTDNet for brevity. Built on top of DiT \cite{peebles2023scalable}, DTDNet leverages a Transformer architecture to implement the diffusion process. For a $512 \times 512$ resolution case, the framework operates as follows: Given a source image containing the desired object $X_s \in \mathbb{R}^{512\times 512\times 3}$ and the target box position $y_b$ within a hint image $H \in \mathbb{R}^{512\times 512\times 3}$ (represented in the VAE latent space as $h \in \mathbb{R}^{64\times 64 \times 4}$), the goal is to learn the model parameters $\theta$ to capture the target box, while preserving the style and content of the object from the source image, to generate the target image $Y \in \mathbb{R}^{512\times 512\times 3}$. To achieve this, we first utilize a pre-trained Stable Diffusion VAE \cite{rombach2022high} to project the pixel-space images into latent representations $x_s \in \mathbb{R}^{64\times 64\times 4}$ and $y \in \mathbb{R}^{64\times 64\times 4}$, facilitating efficient denoising. The denoising network $\epsilon_\theta$, a transformer-based diffusion model \cite{peebles2023scalable}, learns the conditional distribution $p_\theta(y|x_s, y_b)$. During training, Gaussian noise $\epsilon \sim \mathcal{N}(0, \mathbf{I})$ is progressively added to the target image $y$, producing a noisy version $y_t$ at each time step $t \in [1, T]$. The conditions—comprising $x_s$, the box $y_b$, and the hint $h$—are encapsulated in the conditional vector $\vec{c}$. The training objective is to predict the added noise by minimizing the mean squared error:
\begin{equation}
    \mathcal{L_\text{denoising}} = \mathbb{E}_{y,\vec{c},\epsilon\sim\mathcal{N}(0,\mathbf{I}),t}\Vert \epsilon - \epsilon_\theta(y_t,\vec{c},t) \Vert^2.
\end{equation}
After training $p_\theta$, the inference process starts with a random noise image $y_T \sim \mathcal{N}(0, \mathbf{I})$. The model then iteratively samples $y_{t-1} \sim p_\theta(y_{t-1} | y_t)$ until it generates the final target image $y_0$. Our diffusion transformer network is built upon the DiT architecture \cite{peebles2023scalable}. In this process, the noisy target latent $y_t \in \mathbb{R}^{64 \times 64 \times 4}$ is concatenated with the hint latent $h \in \mathbb{R}^{64 \times 64 \times 4}$ to produce $y_{cat} \in \mathbb{R}^{64 \times 64 \times 8}$. This combined latent is divided into patches of size $p = 2$, resulting in a sequence of patches $z_{y_t} = [z_y^{(1)}, z_y^{(2)}, \dots, z_y^{(L_y)}] \in \mathbb{R}^{L_y \times D}$, where $L_y$ denotes the sequence length, and $D$ is the embedding dimension. The conditional vector $\vec{c}$ is integrated into DTDNet through adaptive layer normalization (AdaLN-Zero), following the default settings of the DiT framework. Fig. \ref{fig:spatial_preserved} illustrates examples of training data featuring objects of various sizes and perspectives, often derived from two random frames within a video, as described in \cite{chen2024anydoor}.

\vspace{-6pt}
\subsection{Conditions Collector Network (CCNet)}
The Conditions Collector Network (CCNet) integrates four key inputs: (1) the target box condition feature (TBF), (2) the hint image feature (HIF) extracted from DINOv2, (3) the global source image feature (GSIF) derived from the pre-trained DINOv2 model, and (4) the local source image feature obtained from the VAE. These inputs are processed to generate a compact conditional vector $\vec{c}$, matching the width of the Transformer ($D=1024$ in our model).

\textbf{Local Source Image Feature.}
As discussed in \cite{phamcross}, the local source image feature (LSIF) ensures alignment between the source image and the noisy target image within the DTDNet. This facilitates the transfer of crucial information such as appearance, texture, and identity from the source image to generate a harmonious object within the target box. Specifically, the source image latent ($64 \times 64 \times 4$) is transformed into a sequence of patches $z_{x_s} = [z_x^{(1)}, z_x^{(2)}, \dots, z_x^{(L_x)}] \in \mathbb{R}^{L_x \times D}$, where $L_x = 1024$. This sequence serves as the local source image feature ($\text{LSIF} = z_{x_s} \in \mathbb{R}^{1024 \times D}$).

\textbf{Masked Box Feature.}
The masked box feature is derived from a 3-channel RGB visualization of the target box ($224 \times 224 \times 3$), which serves as input to CCNet. We employ the pre-trained DINOv2-B model to extract features, generating a CLS token and 256 patch tokens that are concatenated into a sequence of 257 tokens ($\text{TBF} \in \mathbb{R}^{257 \times D}$). This sequence is passed through a $1 \times 1$ convolutional layer, reducing the 257 channels to a single channel, resulting in the vector $v_B \in \mathbb{R}^D$.

\textbf{Composed Global Source Image Feature.}
To capture both the identity-preserving details of the object and the global context from the hint image, we extract DINO features from both the source and hint images. Previous works such as AnyDoor \cite{chen2024anydoor} and X-MDPT \cite{phamcross} have demonstrated the effectiveness of self-supervised models like DINO in capturing fine-grained object details and identity. Using DINOv2-G, we extract the CLS token and concatenate it with 256 patch tokens to form the global features ($\text{GSIF1} \in \mathbb{R}^{257 \times D}$ and $\text{GSIF2} \in \mathbb{R}^{257 \times D}$ for the source and hint images, respectively). The input images, originally at $512 \times 512$ resolution, are resized to $224 \times 224$ to comply with DINO's requirements (ensuring both dimensions are divisible by 14). The extracted features $\text{GSIF1}$ and $\text{GSIF2}$ are then combined through a single MLP layer to produce the global source image feature $GSIF \in \mathbb{R}^{514 \times D}$.

\textbf{Compact Conditions Representation.}
To generate the final conditional vector, we concatenate the extracted features ($\text{LSIF}$, $v_B$, $GSIF1$, and $GSIF2$)  along the channel dimension to form $m = 1539$ channels. An $1 \times 1$ convolution operation $\mathcal{C}$ (Fusion) is applied, producing a compact conditional vector $\vec{c} = \mathcal{C}(\text{LSIF}, v_B, \text{GSIF1}, \text{GSIF2}) \in \mathbb{R}^D$. While the hint image provides valuable contextual information for target image prediction, our experiments reveal that combining its VAE and DINO features with those of the source image is essential for accurate generation and faster model convergence. This compact vector significantly reduces the computational load and mitigates the risk of overfitting compared to methods like \cite{chen2024anydoor}, which process the full dimensions of all conditions alongside the noisy target image, leading to higher complexity.

\begin{figure}%[!ht]%[!htbp]
  \centering
  \includegraphics[width=1.0\linewidth]{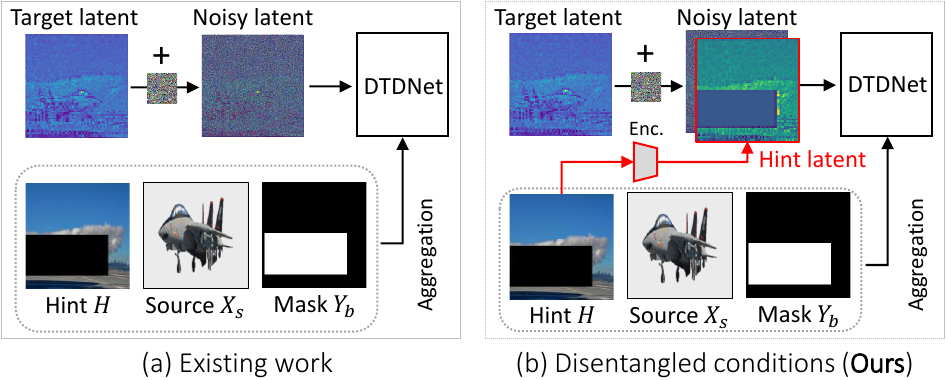}
  \vspace{-16pt}
  \caption{
    \textbf{Existing Design \cite{phamcross} vs. Ours.} Our approach leverages the VAE latent of the hint image to guide noisy target generation, enabling efficient adaptation to visible information outside the boxed region.  
  }
  \label{fig:disentangled}
\vspace{-10pt}
\end{figure}

\subsection{Disentangled Masked Diffusion Module}

\begin{figure}[!ht] %[!htbp]
  \centering
  \includegraphics[width=1\linewidth]{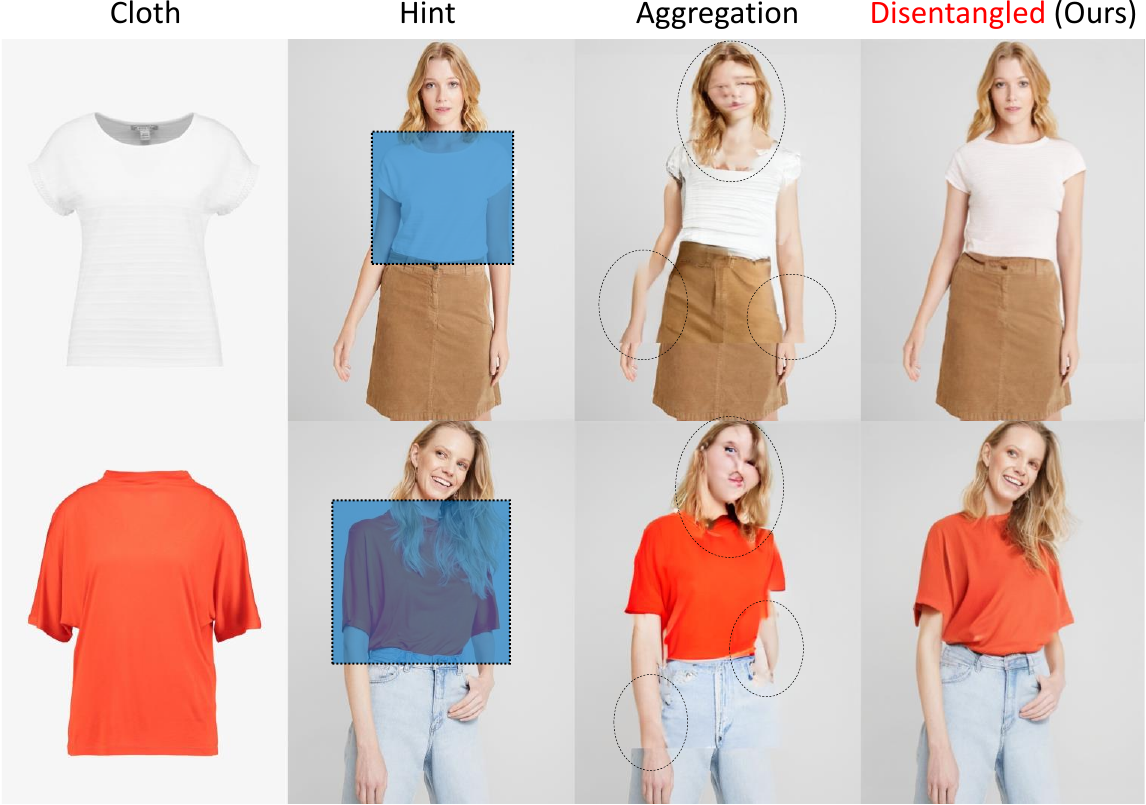}
  \vspace{-14pt}
  \caption{ \textbf{Aggregation vs Disentangled Hinting.} Images are generated by methods on the VITON-HD dataset. Differences are shown more clearly in the face and hand regions. It is best viewed with zoom in at least 200\%. }
  \label{fig:agg_vs_sep}
  \vspace{-8pt}
\end{figure}

Recently, \citet{phamcross} adapted MDT \cite{gao2023masked}, originally designed for class-conditional generation, to human image synthesis by modifying MDT to learn the distribution. Their method combines all conditions into a single vector, which is then used to modulate the noisy target image, achieving strong results in person image generation across various poses. However, for zero-shot object customization, this approach is ineffective (see Fig. \ref{fig:agg_vs_sep}), as the task aligns more closely with conditional inpainting. Here, the target image contains partial information that aids in predicting the missing regions, unlike generating an entire image purely from conditions. Fig. \ref{fig:disentangled} illustrates the distinction between their design and our proposed solution, which we term ``disentangled conditions.'' Our approach separates the conditions into two distinct branches: 
\paragraph{Noisy Latent Processing Branch.}
This branch leverages the VAE hint latent, which is pixel-aligned with the noisy target latent $y_t$, to supervise the reconstruction of the visual information outside the masked region. This design reduces the model's burden to only predicting the missing content within the bounding box. Specifically, we concatenate the hint latent and the noisy target latent along the channel dimension:
\begin{equation}
    y_{\text{cat}}(t) = \text{concat}(h, y_t) \in \mathbb{R}^{64 \times 64 \times 8}.
\end{equation}
The concatenated latent is then patchified using a patch size of $p=2$, following the default configuration of DiT. Unlike previous methods \cite{peebles2023scalable, gao2023masked, phamcross}, which directly patchify the noisy target latent $y_t$, this step better incorporates contextual information. 

\paragraph{Condition Processing Branch.}
The remaining conditions are processed similarly to \citet{phamcross}, yielding a lightweight vector that guides the denoising process. To enhance training, we introduce a masking mechanism in the latent space, randomly masking 30\% of the patchified latent tokens, as in \citet{gao2023masked}. The loss function for training with masked tokens aligns with the standard denoising loss:
\begin{equation}
    \mathcal{L}_{\text{denoising\_mask}} = \mathbb{E}_{y, \vec{c}, \epsilon \sim \mathcal{N}(0, \mathbf{I}), t} \Vert \epsilon - \epsilon_\theta(\mathcal{S}_\theta(x_s, y_m), \vec{c}, t) \Vert^2,
\end{equation}
where $\mathcal{S}_\theta$ is the network comprising $N_1$ encoder layers, a side-interpolator, and $N_2$ decoder layers, as defined in DTDNet. These components, along with other architectural settings, remain consistent with MDT \cite{gao2023masked}. The side-interpolator \cite{gao2023masked} applies self-attention to learn contextual relationships among masked tokens, defined as:
\begin{align}
\begin{split}
    \label{mipnet}
    \mathcal{S}_\theta(z_{y_m}) = z_{y_m} + \varphi_{\text{self-attention}}(z_{y_m}, z_{y_m}, z_{y_m}),
\end{split}
\end{align}
where $\varphi_{\text{self-attention}}$ follows the attention mechanism proposed by \citet{vaswani2017attention}:
\begin{equation}
    \label{attention}
    \varphi_{\text{self-attention}}(\mathbb{Q}, \mathbb{K}, \mathbb{V}) = \text{softmax} \left( \frac{\mathbb{QK}^\top}{\sqrt{d_k}} \right) \mathbb{V}.
\end{equation}

\paragraph{Final Objective Function.}
The model jointly optimizes two loss functions:
\begin{equation}
    \mathcal{L_\text{join}} = \mathcal{L}_{\text{denoising}} + \lambda \mathcal{L}_{\text{denoising\_mask}},
\end{equation}
where the masking branch with the side-interpolator acts as a strong regularizer during training but is omitted during inference, as introduced in MDT \cite{gao2023masked}. This disentangled approach reduces computational complexity and mitigates overfitting, making it more effective for zero-shot object customization. $\lambda$ is set to 1 as MDT's default.

\begin{table*}[!htbp]
    \caption{Comparison of \texttt{E-MD3C} and existing methods on the VITON test set \cite{choi2021viton}, using only bounding boxes without pose or segmentation masks, as required in virtual try-on tasks. All methods are trained on the same data under a zero-shot object customization setting. `$*$' indicates our design with disentangled conditions.}
    \vspace{-10pt}
    \label{tab:compare_deepfashion}
    \begin{center}
    \resizebox{1.0\hsize}{!}{
    \begin{tabular}{ccccccccc}
    \toprule
    \bf  Method &\bf  FID $\downarrow$ &\bf PSNR $\uparrow$ &\bf LPIPS $\downarrow$ &\bf  SSIM $\uparrow$ & \bf L1 $\downarrow$ & \bf Infer. Time (s) $\downarrow$ & \bf \#Params $\downarrow$ & \bf External Data \\
    \midrule
    X-MDPT \cite{phamcross} & 13.70 & 15.75 & 0.2496 & 0.7051 & 8.19E-05 & $\mathbf{2.8\pm 0.02}$ & \bf 468M & \xmark \\
    DiT \cite{peebles2023scalable}$^*$ & 11.78 & 17.84 & 0.1865 & 0.7950 & 5.98E-05 & $\mathbf{2.8\pm 0.02}$ & \bf 468M & \xmark \\
    AnyDoor \cite{chen2024anydoor} & 8.55 & 19.24 & 0.1730 & 0.7992 & 5.56E-05 & 7.1$\pm 0.05$ & 1720M & \cmark \\
    % AnyDoor \cite{chen2024anydoor} & 43.25 & 18.55 & 0.1820 & 0.7857 & 6.11E-05 & 7.1$\pm 0.05$ & 1720M \\
    \midrule
    % \bf E-MD3C (Ours), 850k full & \bf 8.58 & \bf 19.38 & \bf 0.1711 & \bf 0.8080 & \bf 5.18E-05 & $\mathbf{2.8\pm 0.02}$ & \bf 468M \\
    % \bf E-MD3C (Ours), 850k & \bf 44.54 & \bf 18.75 & \bf 0.1785 & \bf 0.7960 & \bf 5.68E-05 & $\mathbf{2.8\pm 0.02}$ & \bf 468M \\
    % \bf E-MD3C (Ours), 950k & \bf 42.54 & \bf 18.48 & \bf 0.1765 & \bf 0.7972 & \bf 5.57E-05 & $\mathbf{2.8\pm 0.02}$ & \bf 468M \\
    % \bf E-MD3C (Ours), 1.1M & \bf 42.92 & \bf 18.62 & \bf 0.1746 & \bf 0.7987 & \bf 5.46E-05 & $\mathbf{2.8\pm 0.02}$ & \bf 468M \\
    % \bf E-MD3C (Ours), 1.1M full & \bf 8.62 & \bf 18.92 & \bf 0.1689 & \bf 0.8097 & \bf 5.05E-05 & $\mathbf{2.8\pm 0.02}$ & \bf 468M \\
    \bf E-MD3C (Ours) & \bf 8.47 & \bf 19.38 & \bf 0.1625 & \bf 0.8106 & \bf 4.92E-05 & $\mathbf{2.8\pm 0.02}$ & \bf 468M & \xmark \\
    \bottomrule
    \end{tabular}}
    \end{center}
    \vspace{-12pt}
\end{table*}

\subsection{Dynamic Classifier-Free Guidance}
We adopt dynamic classifier-free guidance (dy-CFG) \cite{ho2022classifier}, following \cite{gao2023masked, phamcross, pham2025mdsgen}, where the predicted noise is computed as a weighted sum of the unconditional model $\epsilon_\theta(y_t,t)$ and the conditional model $\epsilon_\theta(y_t,\vec{c},t)$: 
\begin{equation}
    \hat{\epsilon}_\theta(y_t,x,t) = \beta_t \epsilon_\theta(y_t,\vec{c},t) + (1-\beta_t)\epsilon_\theta(y_t,t).
\end{equation}  
The guidance scale $\beta_t$ varies dynamically at each timestep $t$. To CFG, we randomly set the conditional vector $\vec{c} \in \mathbb{R}^D$, obtained from CCNet, to a zero vector $\vec{o} \in \mathbb{R}^D$ with a probability of $\eta = 10\%$ during training. The guidance scale follows a power-cosine schedule:  
\begin{equation}
    \beta_t = \frac{1 - \cos \pi (\frac{t}{T})^\gamma}{2} \times \beta,
\end{equation}  
consistent with MDT \cite{gao2023masked}. We use default values of $\beta = 2.0$ and $\gamma = 0.01$.

\section{Experiments}
\label{experiments_sec}
\subsection{Implementation Details}
\paragraph{Dataset.}  
Our training dataset integrates multiple public datasets, including video and multi-view image datasets, to extract source and target images. These are obtained from either two arbitrary video frames or two random object views. We utilize a subset of datasets from AnyDoor \cite{chen2024anydoor}, specifically YouTubeVOS, Saliency, VIPSeg, MVIImageNet, SAM, VITON-HD, Mose, FashionTryon, LVIS, and DressCode, excluding certain datasets due to difficulties in downloading or processing. For evaluation, we focus on high-resolution images from the VITON-HD test set \cite{choi2021viton} at $512 \times 512$ resolution. Preprocessing follows prior work \cite{chen2024anydoor}, with source object images and segmentation masks used DINOv2 \cite{oquab2023dinov2} at $224 \times 224$ input size, while the source object image maintains $512 \times 512$ resolution for input into the VAE.

\vspace{-10pt}
\paragraph{Metrics.}
We adopt standard evaluation metrics from prior research \cite{choi2021viton, bhunia2023person, phamcross}, including FID, PSNR, SSIM, LPIPS, and L1, utilizing evaluation scripts provided by DisCo \cite{wang2024disco}. Our approach is benchmarked against the primary baseline, AnyDoor \cite{chen2024anydoor}, replacing their resource-intensive double Stable Diffusion backbone with our efficient Transformer-based model while maintaining consistent dataset processing for fair comparison. Additionally, for object composition tasks lacking ground truth, we follow prior works by using CLIP and DINO scores to evaluate performance.

\vspace{-10pt}
\paragraph{Training.}
We use the pre-trained VAE with ft-MSE from Stable Diffusion. For $512 \times 512$ images, training was conducted on a single A100 GPU with a batch size 5 for 1.5 million steps. For ablation studies, we trained the model for 300k steps. The learning rate was set to $1 \times 10^{-4}$, with an EMA rate of 0.9999, and other settings were consistent with those in DiT \cite{peebles2023scalable, phamcross}.
\begin{figure*}[!ht] %[!htbp] %
  \centering
  \includegraphics[width=1\linewidth]{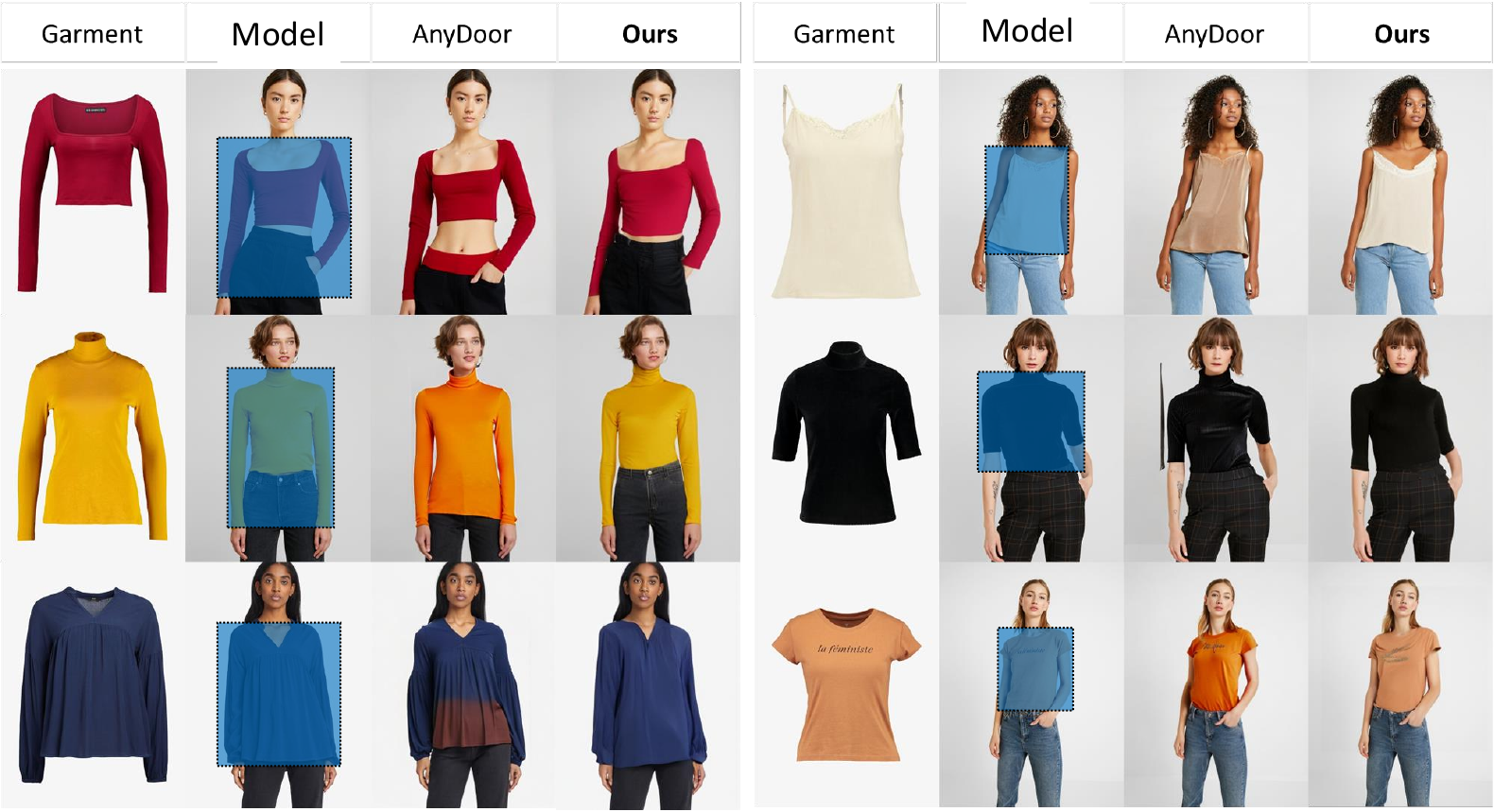} %VITON-HD
  \vspace{-8pt}
  \caption{\textbf{Qualitative Comparison.} Results on the VITON-HD dataset demonstrate the performance of various approaches. Given two inputs $-$ a cloth image and a model hint image with a boxed region indicating where to place the cloth$-$our transformer-based \texttt{E-MD3C} (468M) effectively handles diverse scenarios, producing more realistic and ground truth-aligned outputs compared to the CNN-based method AnyDoor \cite{chen2024anydoor}.}
  \label{fig:vitonhd}
  \vspace{-12pt}
\end{figure*}

\vspace{-6pt}
\begin{table}[!htbp]
    \caption{\textbf{Objects Composition.} Results are presented for image generation at $512\times 512$ resolution on our collected data and DreamBooth \cite{ruiz2023dreambooth} test set.}
    \vspace{-10pt}
    \label{tab:objects_eval}
    \begin{center}
    \resizebox{1.0\hsize}{!}{
    \begin{tabular}{cccc}
    \toprule
    \bf Method & $\text{CLIP}_{score}$ $\uparrow$ & $\text{DINO}_{score}$$\uparrow$ & \bf External Data \\
    \midrule
    AnyDoor \cite{chen2024anydoor}& 0.7306 & \bf 0.4831 & \cmark \\
    \midrule
    \bf E-MD3C (Ours) & \bf 0.7322 & 0.4702 & \xmark \\
    \bottomrule
    \end{tabular}}
    \end{center}
    \vspace{-12pt}
\end{table}

\subsection{Main Results}
\label{deepfashiion_result_sec}
\paragraph{Quantitative Results.}
Tab. \ref{tab:compare_deepfashion} presents a comparative evaluation of various methods for the \textbf{virtual try-on task}, where models operate without pose estimation or segmentation masks, relying only on a clothing item and a rough bounding box. Our \texttt{E-MD3C} consistently outperforms others across FID, SSIM, LPIPS, and L1 metrics at a resolution of $512\times 512$. Unlike AnyDoor \cite{chen2024anydoor}, which depends on a large-scale model pretrained on billions of text-image pairs from the LAION dataset, our method achieves strong results without such extensive pretraining, highlighting its efficiency. \textbf{For object composition} (Tab. \ref{tab:objects_eval}), our method performs comparably to AnyDoor in CLIP scores and slightly lags in DINO scores, likely due to AnyDoor’s backbone being trained on a broader set of objects, including those in DreamBooth datasets. However, the performance gap is small and can be bridged with additional video training data, which is readily available. Given the significantly lower computational demands of our approach, \texttt{E-MD3C} presents a far more practical and scalable solution.

\vspace{-10pt}
\paragraph{Qualitative Results.}
Fig. \ref{fig:vitonhd} compares the outputs of \texttt{E-MD3C} with those of existing methods. Our approach consistently produces high-quality try-on images across various scenarios. In contrast, the Unet-based AnyDoor \cite{chen2024anydoor}, built on a heavy double-Unet ControlNet \cite{zhang2023adding} architecture, often struggles to capture fine-grained clothing details, leading to noticeable artifacts. By leveraging latent-space processing and a transformer with masked modeling for semantic understanding, \texttt{E-MD3C} accurately preserves intricate garment details, resulting in more realistic and visually coherent images. Additional qualitative results are available in the \textbf{Appendix}.

\subsection{Ablation Studies}
\label{alation_sec}
For ablation studies, we evaluate high-resolution VITON-HD \cite{choi2021viton} test images, analyzing metrics and the impact of different configurations. Visual comparisons further illustrate the effectiveness of each modification.

\subsubsection{Impact of Decoupled Conditions}
\begin{figure}[!htbp]
  \centering
  \includegraphics[width=1.0\linewidth]{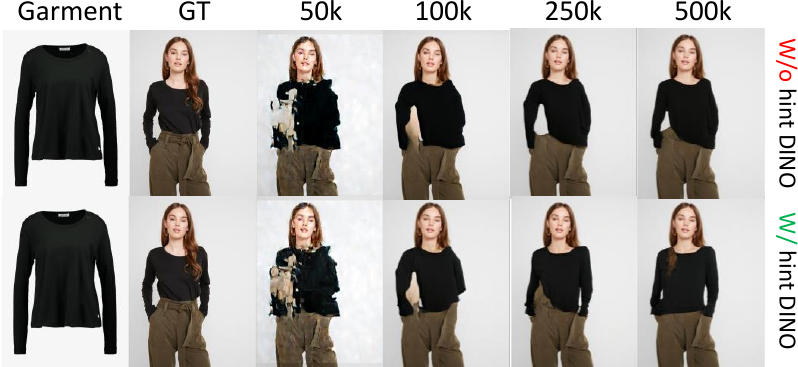}
  \vspace{-10pt}
  \caption{\textbf{Generated images with different training steps}. Adding DINO features of hint image showing a better convergence. It is best viewed with zoom in 200\%.}
  \label{fig:image_steps}
  \vspace{-6pt}
\end{figure}

As shown in Fig. \ref{fig:image_steps}, using only the hint image in the denoising branch alongside the noisy target image fails to achieve optimal convergence. Decoupling the hint image features with DINO and processing them separately in the CCNet condition branch significantly accelerates training and enhances generation accuracy.

\subsubsection{Compared Efficiency}
\label{efficiency}
We quantitatively demonstrate the efficiency and speed advantages of our method over AnyDoor \cite{chen2024anydoor}, as shown in Tab. \ref{tab:compare_memory}. AnyDoor relies on a large, 1720M parameter model that requires 7.1 seconds and 18GB of GPU memory per image generation. In contrast, our lightweight 468M parameter model achieves the same task in just 2.8 seconds while using only 12GB of memory. This significant reduction in both inference time and memory usage highlights the practicality of our Transformer-based framework, especially when compared to the Unet-based Stable Diffusion and the resource-intensive ControlNet-style methods. These results emphasize our method's superior efficiency, making it more practical for real-world applications where computational resources and time are often limited.

\vspace{-6pt}
\begin{table}[!htbp]
    \caption{\textbf{Compare Efficiency.} Results are presented for a single image generation (batch size $\mathcal{B}=1$) for $512\times 512$ resolution, 50 DDIM denoising steps, using one NVIDIA A100 GPU. We conduct 10 runs and take the average.}
    \vspace{-10pt}
    \label{tab:compare_memory}
    \begin{center}
    \resizebox{1.0\hsize}{!}{
    \begin{tabular}{ccccc}
    \toprule
    \bf Method & \bf  Infer. Time$\downarrow$ & \bf Mem. (M)$\downarrow$ & \bf \#Param.$\downarrow$ & \bf Type \\
    \midrule
    AnyDoor [CVPR24] & 7.0 $\pm$ 0.05s & 18273 & 1720M & Unet \\
    \midrule
    \bf E-MD3C (Ours) & \bf 2.8 $\pm$ 0.02s & \bf 12071 & \bf 468M & \bf Trans.  \\
    \end{tabular}}
    \end{center}
    \vspace{-10pt}
\end{table}

\subsubsection{Hint Image as Supervision}
\label{hint_image}
Fig. \ref{fig:agg_vs_sep} highlights the limitations of the aggregation-based approach \cite{phamcross}, which fails in zero-shot object customization. Unlike pose-guided person generation, where source and target images share similar structures, target images in object customization often contain unseen parts and backgrounds. This emphasizes the need for a more flexible, disentangled condition representation.

\subsubsection{Impact of Masking Modeling}
\label{impact_mask}
\begin{figure}%[!htbp]
  \centering
  \includegraphics[width=1.0\linewidth]{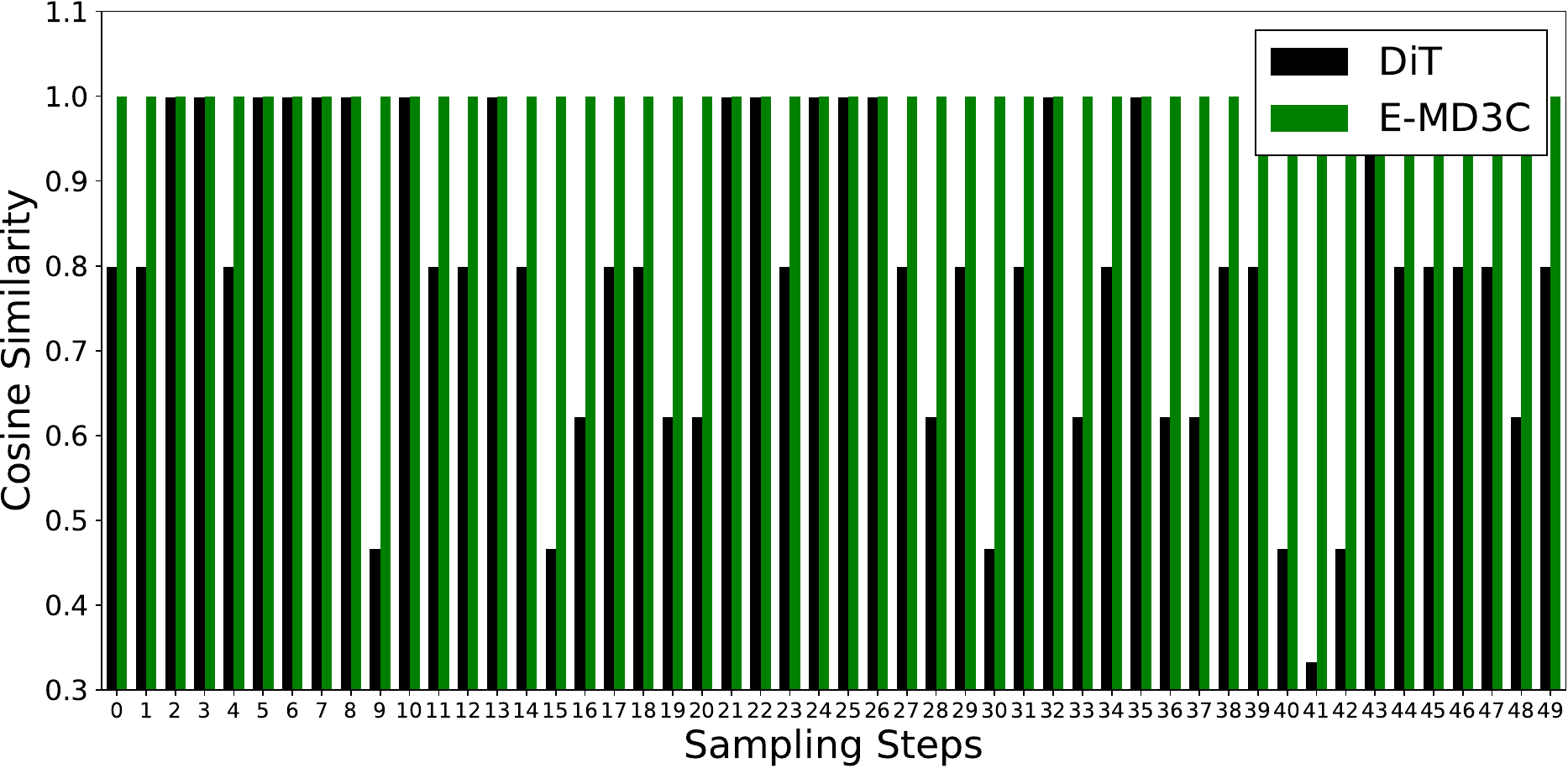} %conditions_similarity
  \vspace{-16pt}
  \caption{Conditional alignment of various source object image views. Our transformer-based \texttt{E-MD3C} model achieves superior alignment scores, measured via cosine similarity, compared to baseline foundation models.}
  \label{fig:cosine}
  \vspace{-6pt}
\end{figure}

% We analyze the impact of masking learning in the zero-shot object customization task, particularly in the multiview learning objective. To evaluate this, we examine how the learnable vector from CCNet correlates with different views of an image in the MVImageNet \cite{yu2023mvimgnet} test set. Our findings reveal that without masking modeling, DiT \cite{peebles2023scalable} struggles to maintain high similarity between views (see Fig. \ref{fig:cosine}). In contrast, our method, equipped with masking learning, achieves significantly higher view consistency, indicating that the model effectively captures essential object features, ensuring robustness across diverse perspectives.
We analyze the impact of masking learning on the zero-shot object customization task, particularly in multiview learning. By examining how the learnable vector from CCNet correlates with different views in the MVImageNet \cite{yu2023mvimgnet} test set, we find that DiT \cite{peebles2023scalable} struggles to maintain view consistency without masking modeling (see Fig. \ref{fig:cosine}). In contrast, our method with masking learning achieves significantly higher similarity across views, demonstrating that it effectively captures key object features and remains robust across various perspectives.

\vspace{-6pt}
\subsubsection{Distribution Statistic}
\label{sec:histogram}
\begin{figure}%[!htbp]
  \centering
  \includegraphics[width=1.0\linewidth]{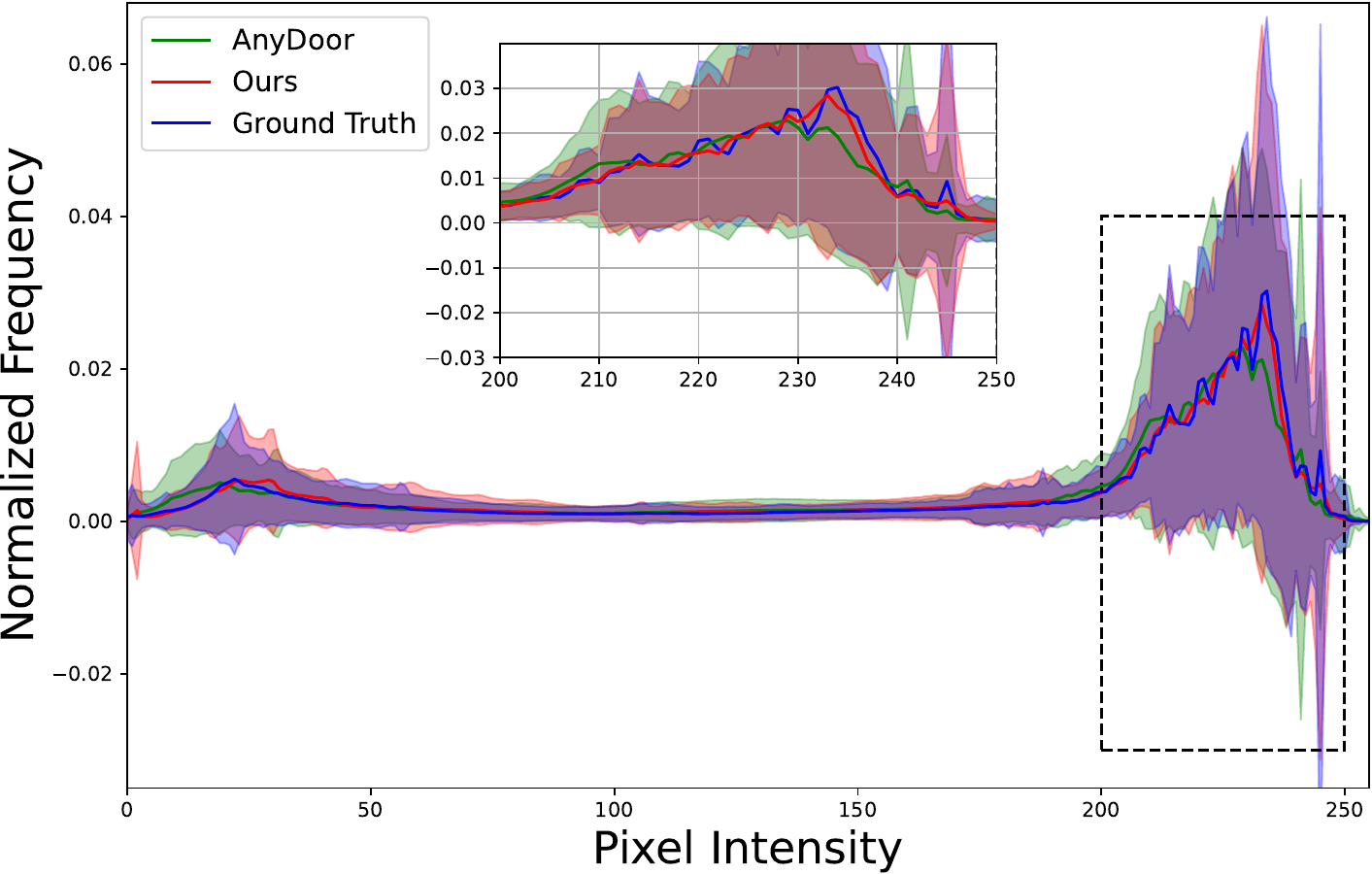}
  \vspace{-16pt}
  \caption{\textbf{Pixel Distribution.} Compare generated images and ground truth (GT). Our method aligns more closely with the GT curve, while AnyDoor \cite{chen2024anydoor} exhibits noticeable deviations. Best viewed at 200\% zoom.}
  \label{fig:histogram}
  \vspace{-10pt}
\end{figure}

To compare the behavior of our model with existing methods, we compute the pixel statistics of the outputs generated by both approaches on the VITON-HD dataset. As shown in Fig. \ref{fig:histogram}, our method produces a pixel distribution that aligns more closely with the ground truth compared to the AnyDoor model. This suggests that the Transformer-based diffusion model captures the data distribution more effectively than the Unet-based model.

\vspace{-8pt}
\section{Conclusion}
\label{conclusions_sec}
In this paper, we present E-MD3C, a novel masked diffusion generative model tailored for zero-shot object customization (ZSOIC). Unlike previous methods that rely on resource-heavy Unet-based backbones for denoising diffusion, our approach utilizes a masked diffusion transformer architecture operating on latent patches, significantly improving efficiency. Extensive experiments demonstrate that E-MD3C not only produces high-quality, high-resolution images but also achieves faster inference speeds, highlighting its potential as an effective and efficient solution for this task.

\newpage
\section*{Impact Statement}
Our method effectively generates high-quality images across various object customization tasks, offering a flexible virtual try-on solution with just a simple bounding box. It provides significant advantages, including faster image generation compared to existing large, slow models. However, as with all image synthesis technologies, there is a potential for misuse, such as creating deceptive content. We are committed to implementing safeguards to regulate access, ensuring the technology benefits the community while minimizing risks.

\nocite{pham2023self}
\nocite{pham2022lad}
\nocite{pham2022pros}

\bibliography{conferences}

\begin{thebibliography}{65}
\providecommand{\natexlab}[1]{#1}
\providecommand{\url}[1]{\texttt{#1}}
\expandafter\ifx\csname urlstyle\endcsname\relax
  \providecommand{\doi}[1]{doi: #1}\else
  \providecommand{\doi}{doi: \begingroup \urlstyle{rm}\Url}\fi

\bibitem[Athar et~al.(2023)Athar, Luiten, Voigtlaender, Khurana, Dave, Leibe,
  and Ramanan]{athar2023burst}
Athar, A., Luiten, J., Voigtlaender, P., Khurana, T., Dave, A., Leibe, B., and
  Ramanan, D.
\newblock Burst: A benchmark for unifying object recognition, segmentation and
  tracking in video.
\newblock In \emph{Proceedings of the IEEE/CVF winter conference on
  applications of computer vision}, pp.\  1674--1683, 2023.

\bibitem[Bhunia et~al.(2023)Bhunia, Khan, Cholakkal, Anwer, Laaksonen, Shah,
  and Khan]{bhunia2023person}
Bhunia, A.~K., Khan, S., Cholakkal, H., Anwer, R.~M., Laaksonen, J., Shah, M.,
  and Khan, F.~S.
\newblock Person image synthesis via denoising diffusion model.
\newblock In \emph{Proceedings of the IEEE/CVF Conference on Computer Vision
  and Pattern Recognition}, pp.\  5968--5976, 2023.

\bibitem[Borji et~al.(2015)Borji, Cheng, Jiang, and Li]{borji2015salient}
Borji, A., Cheng, M.-M., Jiang, H., and Li, J.
\newblock Salient object detection: A benchmark.
\newblock \emph{IEEE transactions on image processing}, 24\penalty0
  (12):\penalty0 5706--5722, 2015.

\bibitem[Chang et~al.(2022)Chang, Zhang, Jiang, Liu, and
  Freeman]{chang2022maskgit}
Chang, H., Zhang, H., Jiang, L., Liu, C., and Freeman, W.~T.
\newblock Maskgit: Masked generative image transformer.
\newblock In \emph{Proceedings of the IEEE/CVF Conference on Computer Vision
  and Pattern Recognition}, pp.\  11315--11325, 2022.

\bibitem[Chang et~al.(2023)Chang, Zhang, Barber, Maschinot, Lezama, Jiang,
  Yang, Murphy, Freeman, Rubinstein, et~al.]{chang2023muse}
Chang, H., Zhang, H., Barber, J., Maschinot, A., Lezama, J., Jiang, L., Yang,
  M.-H., Murphy, K., Freeman, W.~T., Rubinstein, M., et~al.
\newblock Muse: Text-to-image generation via masked generative transformers.
\newblock \emph{arXiv preprint arXiv:2301.00704}, 2023.

\bibitem[Chen et~al.(2025)Chen, Ge, Xie, Wu, Yao, Ren, Wang, Luo, Lu, and
  Li]{chen2025pixart}
Chen, J., Ge, C., Xie, E., Wu, Y., Yao, L., Ren, X., Wang, Z., Luo, P., Lu, H.,
  and Li, Z.
\newblock Pixart: Weak-to-strong training of diffusion transformer for 4k
  text-to-image generation.
\newblock In \emph{European Conference on Computer Vision}, pp.\  74--91.
  Springer, 2025.

\bibitem[Chen et~al.(2024{\natexlab{a}})Chen, Feng, Chen, Wang, Zhang, Liu,
  Shen, and Zhao]{chen2024zeroshot}
Chen, X., Feng, Y., Chen, M., Wang, Y., Zhang, S., Liu, Y., Shen, Y., and Zhao,
  H.
\newblock Zero-shot image editing with reference imitation.
\newblock In \emph{The Thirty-eighth Annual Conference on Neural Information
  Processing Systems}, 2024{\natexlab{a}}.
\newblock URL \url{https://openreview.net/forum?id=LZV0U6UHb6}.

\bibitem[Chen et~al.(2024{\natexlab{b}})Chen, Huang, Liu, Shen, Zhao, and
  Zhao]{chen2024anydoor}
Chen, X., Huang, L., Liu, Y., Shen, Y., Zhao, D., and Zhao, H.
\newblock Anydoor: Zero-shot object-level image customization.
\newblock In \emph{Proceedings of the IEEE/CVF Conference on Computer Vision
  and Pattern Recognition}, pp.\  6593--6602, 2024{\natexlab{b}}.

\bibitem[Choi et~al.(2021)Choi, Park, Lee, and Choo]{choi2021viton}
Choi, S., Park, S., Lee, M., and Choo, J.
\newblock Viton-hd: High-resolution virtual try-on via misalignment-aware
  normalization.
\newblock In \emph{Proceedings of the IEEE/CVF conference on computer vision
  and pattern recognition}, pp.\  14131--14140, 2021.

\bibitem[Cong et~al.(2020)Cong, Zhang, Niu, Liu, Ling, Li, and
  Zhang]{cong2020dovenet}
Cong, W., Zhang, J., Niu, L., Liu, L., Ling, Z., Li, W., and Zhang, L.
\newblock Dovenet: Deep image harmonization via domain verification.
\newblock In \emph{Proceedings of the IEEE/CVF conference on computer vision
  and pattern recognition}, pp.\  8394--8403, 2020.

\bibitem[Devlin et~al.(2018)Devlin, Chang, Lee, and Toutanova]{devlin2018bert}
Devlin, J., Chang, M.-W., Lee, K., and Toutanova, K.
\newblock Bert: Pre-training of deep bidirectional transformers for language
  understanding.
\newblock \emph{arXiv preprint arXiv:1810.04805}, 2018.

\bibitem[Ding et~al.(2023)Ding, Liu, He, Jiang, Torr, and Bai]{ding2023mose}
Ding, H., Liu, C., He, S., Jiang, X., Torr, P.~H., and Bai, S.
\newblock Mose: A new dataset for video object segmentation in complex scenes.
\newblock In \emph{Proceedings of the IEEE/CVF International Conference on
  Computer Vision}, pp.\  20224--20234, 2023.

\bibitem[Dosovitskiy et~al.(2010)Dosovitskiy, Beyer, Kolesnikov, Weissenborn,
  Zhai, Unterthiner, Dehghani, Minderer, Heigold, Gelly,
  et~al.]{dosovitskiy2010image}
Dosovitskiy, A., Beyer, L., Kolesnikov, A., Weissenborn, D., Zhai, X.,
  Unterthiner, T., Dehghani, M., Minderer, M., Heigold, G., Gelly, S., et~al.
\newblock An image is worth 16x16 words: Transformers for image recognition at
  scale. arxiv 2020.
\newblock \emph{arXiv preprint arXiv:2010.11929}, 2010.

\bibitem[Dosovitskiy et~al.(2021)Dosovitskiy, Beyer, Kolesnikov, Weissenborn,
  Zhai, Unterthiner, Dehghani, Minderer, Heigold, Gelly, Uszkoreit, and
  Houlsby]{dosovitskiy2021an}
Dosovitskiy, A., Beyer, L., Kolesnikov, A., Weissenborn, D., Zhai, X.,
  Unterthiner, T., Dehghani, M., Minderer, M., Heigold, G., Gelly, S.,
  Uszkoreit, J., and Houlsby, N.
\newblock An image is worth 16x16 words: Transformers for image recognition at
  scale.
\newblock In \emph{ICLR}, 2021.

\bibitem[Esser et~al.(2024)Esser, Kulal, Blattmann, Entezari, M{\"u}ller,
  Saini, Levi, Lorenz, Sauer, Boesel, et~al.]{esser2024scaling}
Esser, P., Kulal, S., Blattmann, A., Entezari, R., M{\"u}ller, J., Saini, H.,
  Levi, Y., Lorenz, D., Sauer, A., Boesel, F., et~al.
\newblock Scaling rectified flow transformers for high-resolution image
  synthesis.
\newblock In \emph{Forty-first International Conference on Machine Learning},
  2024.

\bibitem[Gal et~al.(2022)Gal, Alaluf, Atzmon, Patashnik, Bermano, Chechik, and
  Cohen-Or]{gal2022image}
Gal, R., Alaluf, Y., Atzmon, Y., Patashnik, O., Bermano, A.~H., Chechik, G.,
  and Cohen-Or, D.
\newblock An image is worth one word: Personalizing text-to-image generation
  using textual inversion.
\newblock \emph{arXiv preprint arXiv:2208.01618}, 2022.

\bibitem[Gao et~al.(2023)Gao, Zhou, Cheng, and Yan]{gao2023masked}
Gao, S., Zhou, P., Cheng, M.-M., and Yan, S.
\newblock Masked diffusion transformer is a strong image synthesizer.
\newblock In \emph{Proceedings of the IEEE/CVF International Conference on
  Computer Vision}, pp.\  23164--23173, 2023.

\bibitem[Ginesu et~al.(2012)Ginesu, Pintus, and Giusto]{ginesu2012objective}
Ginesu, G., Pintus, M., and Giusto, D.~D.
\newblock Objective assessment of the webp image coding algorithm.
\newblock \emph{Signal Processing: Image Communication}, 2012.

\bibitem[Gupta et~al.(2019)Gupta, Dollar, and Girshick]{gupta2019lvis}
Gupta, A., Dollar, P., and Girshick, R.
\newblock Lvis: A dataset for large vocabulary instance segmentation.
\newblock In \emph{Proceedings of the IEEE/CVF conference on computer vision
  and pattern recognition}, pp.\  5356--5364, 2019.

\bibitem[He et~al.(2022)He, Chen, Xie, Li, Doll{\'a}r, and
  Girshick]{he2022masked}
He, K., Chen, X., Xie, S., Li, Y., Doll{\'a}r, P., and Girshick, R.
\newblock Masked autoencoders are scalable vision learners.
\newblock In \emph{Proceedings of the IEEE/CVF conference on computer vision
  and pattern recognition}, pp.\  16000--16009, 2022.

\bibitem[Ho \& Salimans(2022)Ho and Salimans]{ho2022classifier}
Ho, J. and Salimans, T.
\newblock Classifier-free diffusion guidance.
\newblock \emph{arXiv preprint arXiv:2207.12598}, 2022.

\bibitem[Ho et~al.(2020)Ho, Jain, and Abbeel]{ho2020denoising}
Ho, J., Jain, A., and Abbeel, P.
\newblock Denoising diffusion probabilistic models.
\newblock \emph{Advances in neural information processing systems},
  33:\penalty0 6840--6851, 2020.

\bibitem[Ho et~al.(2022)Ho, Saharia, Chan, Fleet, Norouzi, and
  Salimans]{ho2022cascaded}
Ho, J., Saharia, C., Chan, W., Fleet, D.~J., Norouzi, M., and Salimans, T.
\newblock Cascaded diffusion models for high fidelity image generation.
\newblock \emph{The Journal of Machine Learning Research}, 23\penalty0
  (1):\penalty0 2249--2281, 2022.

\bibitem[Jin et~al.(2025)Jin, Shen, Zhao, Fu, and Yang]{jin2025unicanvas}
Jin, J., Shen, Y., Zhao, X., Fu, Z., and Yang, J.
\newblock Unicanvas: Affordance-aware unified real image editing via customized
  text-to-image generation.
\newblock \emph{International Journal of Computer Vision}, pp.\  1--25, 2025.

\bibitem[Jung et~al.(2020)Jung, Hong, Wang, Han, Pham, Park, Kim, Kang, Yoo,
  and Lee]{jung2020flexible}
Jung, Y.~H., Hong, S.~K., Wang, H.~S., Han, J.~H., Pham, T.~X., Park, H., Kim,
  J., Kang, S., Yoo, C.~D., and Lee, K.~J.
\newblock Flexible piezoelectric acoustic sensors and machine learning for
  speech processing.
\newblock \emph{Advanced Materials}, 32\penalty0 (35):\penalty0 1904020, 2020.

\bibitem[Jung et~al.(2022)Jung, Pham, Issa, Wang, Lee, Chung, Lee, Kim, Yoo,
  and Lee]{jung2022deep}
Jung, Y.~H., Pham, T.~X., Issa, D., Wang, H.~S., Lee, J.~H., Chung, M., Lee,
  B.-Y., Kim, G., Yoo, C.~D., and Lee, K.~J.
\newblock Deep learning-based noise robust flexible piezoelectric acoustic
  sensors for speech processing.
\newblock \emph{Nano Energy}, 101:\penalty0 107610, 2022.

\bibitem[Kim et~al.(2020)Kim, Ma, Pham, Kim, and Yoo]{kim2020modality}
Kim, J., Ma, M., Pham, T., Kim, K., and Yoo, C.~D.
\newblock Modality shifting attention network for multi-modal video question
  answering.
\newblock In \emph{Proceedings of the IEEE/CVF conference on computer vision
  and pattern recognition}, pp.\  10106--10115, 2020.

\bibitem[Kirillov et~al.(2023)Kirillov, Mintun, Ravi, Mao, Rolland, Gustafson,
  Xiao, Whitehead, Berg, Lo, et~al.]{kirillov2023segment}
Kirillov, A., Mintun, E., Ravi, N., Mao, H., Rolland, C., Gustafson, L., Xiao,
  T., Whitehead, S., Berg, A.~C., Lo, W.-Y., et~al.
\newblock Segment anything.
\newblock In \emph{Proceedings of the IEEE/CVF International Conference on
  Computer Vision}, pp.\  4015--4026, 2023.

\bibitem[Lee et~al.(2020)Lee, Park, Pham, and Yoo]{lee2020learning}
Lee, D., Park, H., Pham, T., and Yoo, C.~D.
\newblock Learning augmentation network via influence functions.
\newblock In \emph{Proceedings of the IEEE/CVF Conference on Computer Vision
  and Pattern Recognition}, pp.\  10961--10970, 2020.

\bibitem[Li et~al.(2024)Li, Wang, Liu, Du, Sun, Guo, Zhang, and
  Jiang]{li2024quality}
Li, C., Wang, R., Liu, L., Du, J., Sun, Y., Guo, Z., Zhang, Z., and Jiang, Y.
\newblock Quality-aware masked diffusion transformer for enhanced music
  generation.
\newblock \emph{arXiv preprint arXiv:2405.15863}, 2024.

\bibitem[Mao et~al.(2024)Mao, Jiang, Wang, Fu, Zhang, Wu, Wang, Wang, Li, and
  Chi]{mao2024mdt}
Mao, X., Jiang, Z., Wang, Q., Fu, C., Zhang, J., Wu, J., Wang, Y., Wang, C.,
  Li, W., and Chi, M.
\newblock Mdt-a2g: Exploring masked diffusion transformers for co-speech
  gesture generation.
\newblock \emph{arXiv preprint arXiv:2408.03312}, 2024.

\bibitem[Miao et~al.(2022)Miao, Wang, Wu, Li, Zhang, Wei, and
  Yang]{miao2022large}
Miao, J., Wang, X., Wu, Y., Li, W., Zhang, X., Wei, Y., and Yang, Y.
\newblock Large-scale video panoptic segmentation in the wild: A benchmark.
\newblock In \emph{Proceedings of the IEEE/CVF Conference on Computer Vision
  and Pattern Recognition}, pp.\  21033--21043, 2022.

\bibitem[Niu et~al.(2023)Niu, Zhang, Pham, Sun, Zhu, Kweon, and
  Zhang]{niu2023cdpmsr}
Niu, A., Zhang, K., Pham, T.~X., Sun, J., Zhu, Y., Kweon, I.~S., and Zhang, Y.
\newblock Cdpmsr: Conditional diffusion probabilistic models for single image
  super-resolution.
\newblock In \emph{2023 IEEE International Conference on Image Processing
  (ICIP)}, pp.\  615--619. IEEE, 2023.

\bibitem[Niu et~al.(2024{\natexlab{a}})Niu, Pham, Zhang, Sun, Zhu, Yan, Kweon,
  and Zhang]{niu2024acdmsr}
Niu, A., Pham, T.~X., Zhang, K., Sun, J., Zhu, Y., Yan, Q., Kweon, I.~S., and
  Zhang, Y.
\newblock Acdmsr: Accelerated conditional diffusion models for single image
  super-resolution.
\newblock \emph{IEEE Transactions on Broadcasting}, 2024{\natexlab{a}}.

\bibitem[Niu et~al.(2024{\natexlab{b}})Niu, Zhang, Pham, Wang, Sun, Kweon, and
  Zhang]{niu2024learning}
Niu, A., Zhang, K., Pham, T.~X., Wang, P., Sun, J., Kweon, I.~S., and Zhang, Y.
\newblock Learning from multi-perception features for real-word image
  super-resolution.
\newblock \emph{IEEE Transactions on Circuits and Systems for Video
  Technology}, 2024{\natexlab{b}}.

\bibitem[Oquab et~al.(2023)Oquab, Darcet, Moutakanni, Vo, Szafraniec, Khalidov,
  Fernandez, Haziza, Massa, El-Nouby, Howes, Huang, Xu, Sharma, Li, Galuba,
  Rabbat, Assran, Ballas, Synnaeve, Misra, Jegou, Mairal, Labatut, Joulin, and
  Bojanowski]{oquab2023dinov2}
Oquab, M., Darcet, T., Moutakanni, T., Vo, H.~V., Szafraniec, M., Khalidov, V.,
  Fernandez, P., Haziza, D., Massa, F., El-Nouby, A., Howes, R., Huang, P.-Y.,
  Xu, H., Sharma, V., Li, S.-W., Galuba, W., Rabbat, M., Assran, M., Ballas,
  N., Synnaeve, G., Misra, I., Jegou, H., Mairal, J., Labatut, P., Joulin, A.,
  and Bojanowski, P.
\newblock Dinov2: Learning robust visual features without supervision, 2023.

\bibitem[Peebles \& Xie(2023)Peebles and Xie]{peebles2023scalable}
Peebles, W. and Xie, S.
\newblock Scalable diffusion models with transformers.
\newblock In \emph{Proceedings of the IEEE/CVF International Conference on
  Computer Vision}, pp.\  4195--4205, 2023.

\bibitem[Pham et~al.(2022{\natexlab{a}})Pham, Zhang, Niu, Zhang, and
  Yoo]{pham2022pros}
Pham, T., Zhang, C., Niu, A., Zhang, K., and Yoo, C.~D.
\newblock On the pros and cons of momentum encoder in self-supervised visual
  representation learning.
\newblock \emph{arXiv preprint arXiv:2208.05744}, 2022{\natexlab{a}}.

\bibitem[Pham et~al.(2021)Pham, Mina, Issa, and Yoo]{pham2021self}
Pham, T.~X., Mina, R. J.~L., Issa, D., and Yoo, C.~D.
\newblock Self-supervised learning with local attention-aware feature.
\newblock \emph{arXiv preprint arXiv:2108.00475}, 2021.

\bibitem[Pham et~al.(2022{\natexlab{b}})Pham, Mina, Nguyen, Madjid, Choi, and
  Yoo]{pham2022lad}
Pham, T.~X., Mina, R. J.~L., Nguyen, T., Madjid, S.~R., Choi, J., and Yoo,
  C.~D.
\newblock Lad: A hybrid deep learning system for benign paroxysmal positional
  vertigo disorders diagnostic.
\newblock \emph{IEEE Access}, 2022{\natexlab{b}}.

\bibitem[Pham et~al.(2023)Pham, Niu, Zhang, Jin, Hong, and Yoo]{pham2023self}
Pham, T.~X., Niu, A., Zhang, K., Jin, T. J.~T., Hong, J.~W., and Yoo, C.~D.
\newblock Self-supervised visual representation learning via residual momentum.
\newblock \emph{IEEE Access}, 2023.

\bibitem[Pham et~al.(2024)Pham, Zhang, and Yoo]{phamcross}
Pham, T.~X., Zhang, K., and Yoo, C.~D.
\newblock Cross-view masked diffusion transformers for person image synthesis.
\newblock In \emph{Forty-first International Conference on Machine Learning},
  2024.

\bibitem[Pham et~al.(2025)Pham, Ton, and Yoo]{pham2025mdsgen}
Pham, T.~X., Ton, T., and Yoo, C.~D.
\newblock {MDSG}en: Fast and efficient masked diffusion temporal-aware
  transformers for open-domain sound generation.
\newblock In \emph{International Conference on Learning Representations}, 2025.
\newblock URL \url{https://openreview.net/forum?id=yFEqYwgttJ}.

\bibitem[Rombach et~al.(2022)Rombach, Blattmann, Lorenz, Esser, and
  Ommer]{rombach2022high}
Rombach, R., Blattmann, A., Lorenz, D., Esser, P., and Ommer, B.
\newblock High-resolution image synthesis with latent diffusion models.
\newblock In \emph{Proceedings of the IEEE/CVF conference on computer vision
  and pattern recognition}, pp.\  10684--10695, 2022.

\bibitem[Ronneberger et~al.(2015)Ronneberger, Fischer, and
  Brox]{ronneberger2015u}
Ronneberger, O., Fischer, P., and Brox, T.
\newblock U-net: Convolutional networks for biomedical image segmentation.
\newblock In \emph{International Conference on Medical image computing and
  computer-assisted intervention}, 2015.

\bibitem[Ruiz et~al.(2023)Ruiz, Li, Jampani, Pritch, Rubinstein, and
  Aberman]{ruiz2023dreambooth}
Ruiz, N., Li, Y., Jampani, V., Pritch, Y., Rubinstein, M., and Aberman, K.
\newblock Dreambooth: Fine tuning text-to-image diffusion models for
  subject-driven generation.
\newblock In \emph{Proceedings of the IEEE/CVF conference on computer vision
  and pattern recognition}, pp.\  22500--22510, 2023.

\bibitem[Song et~al.(2020)Song, Meng, and Ermon]{song2020denoising}
Song, J., Meng, C., and Ermon, S.
\newblock Denoising diffusion implicit models.
\newblock \emph{arXiv preprint arXiv:2010.02502}, 2020.

\bibitem[Song \& Ermon(2019)Song and Ermon]{song2019generative}
Song, Y. and Ermon, S.
\newblock Generative modeling by estimating gradients of the data distribution.
\newblock \emph{Advances in neural information processing systems}, 32, 2019.

\bibitem[Song et~al.(2023)Song, Zhang, Lin, Cohen, Price, Zhang, Kim, and
  Aliaga]{song2023objectstitch}
Song, Y., Zhang, Z., Lin, Z., Cohen, S., Price, B., Zhang, J., Kim, S.~Y., and
  Aliaga, D.
\newblock Objectstitch: Object compositing with diffusion model.
\newblock In \emph{Proceedings of the IEEE/CVF Conference on Computer Vision
  and Pattern Recognition}, pp.\  18310--18319, 2023.

\bibitem[Trung \& Yoo(2019)Trung and Yoo]{trungshort}
Trung, P.~X. and Yoo, C.~D.
\newblock Short convolutional neural network and mfccs for accurate speaker
  recognition systems.
\newblock \emph{International Technical Conference on Circuits/Systems,
  Computers and Communications (ITC-CSCC)}, 2019.

\bibitem[Vaswani et~al.(2017)Vaswani, Shazeer, Parmar, Uszkoreit, Jones, Gomez,
  Kaiser, and Polosukhin]{vaswani2017attention}
Vaswani, A., Shazeer, N., Parmar, N., Uszkoreit, J., Jones, L., Gomez, A.~N.,
  Kaiser, L., and Polosukhin, I.
\newblock Attention is all you need.
\newblock In \emph{NeurIPS}, 2017.

\bibitem[Vu et~al.(2019)Vu, Jang, Pham, and Yoo]{vu2019cascade}
Vu, T., Jang, H., Pham, T.~X., and Yoo, C.
\newblock Cascade rpn: Delving into high-quality region proposal network with
  adaptive convolution.
\newblock \emph{Advances in neural information processing systems}, 32, 2019.

\bibitem[Wang et~al.(2017)Wang, Lu, Wang, Feng, Wang, Yin, and
  Ruan]{wang2017learning}
Wang, L., Lu, H., Wang, Y., Feng, M., Wang, D., Yin, B., and Ruan, X.
\newblock Learning to detect salient objects with image-level supervision.
\newblock In \emph{Proceedings of the IEEE conference on computer vision and
  pattern recognition}, pp.\  136--145, 2017.

\bibitem[Wang et~al.(2024)Wang, Li, Lin, Zhai, Lin, Yang, Zhang, Liu, and
  Wang]{wang2024disco}
Wang, T., Li, L., Lin, K., Zhai, Y., Lin, C.-C., Yang, Z., Zhang, H., Liu, Z.,
  and Wang, L.
\newblock Disco: Disentangled control for realistic human dance generation.
\newblock In \emph{Proceedings of the IEEE/CVF Conference on Computer Vision
  and Pattern Recognition}, pp.\  9326--9336, 2024.

\bibitem[Wang et~al.(2021)Wang, Feiszli, Wang, and Tran]{wang2021unidentified}
Wang, W., Feiszli, M., Wang, H., and Tran, D.
\newblock Unidentified video objects: A benchmark for dense, open-world
  segmentation.
\newblock In \emph{Proceedings of the IEEE/CVF international conference on
  computer vision}, pp.\  10776--10785, 2021.

\bibitem[Xie et~al.(2025)Xie, Chen, Chen, Cai, Tang, Lin, Zhang, Li, Zhu, Lu,
  et~al.]{xie2025sana}
Xie, E., Chen, J., Chen, J., Cai, H., Tang, H., Lin, Y., Zhang, Z., Li, M.,
  Zhu, L., Lu, Y., et~al.
\newblock {SANA}: Efficient high-resolution text-to-image synthesis with linear
  diffusion transformers.
\newblock In \emph{The Thirteenth International Conference on Learning
  Representations}, 2025.
\newblock URL \url{https://openreview.net/forum?id=N8Oj1XhtYZ}.

\bibitem[Xu et~al.(2018)Xu, Yang, Fan, Yue, Liang, Yang, and
  Huang]{xu2018youtube}
Xu, N., Yang, L., Fan, Y., Yue, D., Liang, Y., Yang, J., and Huang, T.
\newblock Youtube-vos: A large-scale video object segmentation benchmark.
\newblock \emph{arXiv preprint arXiv:1809.03327}, 2018.

\bibitem[Yang et~al.(2023)Yang, Gu, Zhang, Zhang, Chen, Sun, Chen, and
  Wen]{yang2023paint}
Yang, B., Gu, S., Zhang, B., Zhang, T., Chen, X., Sun, X., Chen, D., and Wen,
  F.
\newblock Paint by example: Exemplar-based image editing with diffusion models.
\newblock In \emph{Proceedings of the IEEE/CVF Conference on Computer Vision
  and Pattern Recognition}, pp.\  18381--18391, 2023.

\bibitem[Yang et~al.(2019)Yang, Fan, and Xu]{yang2019video}
Yang, L., Fan, Y., and Xu, N.
\newblock Video instance segmentation.
\newblock In \emph{Proceedings of the IEEE/CVF international conference on
  computer vision}, pp.\  5188--5197, 2019.

\bibitem[Yu et~al.(2023{\natexlab{a}})Yu, Cheng, Sohn, Lezama, Zhang, Chang,
  Hauptmann, Yang, Hao, Essa, et~al.]{yu2023magvit}
Yu, L., Cheng, Y., Sohn, K., Lezama, J., Zhang, H., Chang, H., Hauptmann,
  A.~G., Yang, M.-H., Hao, Y., Essa, I., et~al.
\newblock Magvit: Masked generative video transformer.
\newblock In \emph{Proceedings of the IEEE/CVF Conference on Computer Vision
  and Pattern Recognition}, pp.\  10459--10469, 2023{\natexlab{a}}.

\bibitem[Yu et~al.(2023{\natexlab{b}})Yu, Xu, Zhang, Liu, Ye, Wu, Yan, Zhu,
  Xiong, Liang, et~al.]{yu2023mvimgnet}
Yu, X., Xu, M., Zhang, Y., Liu, H., Ye, C., Wu, Y., Yan, Z., Zhu, C., Xiong,
  Z., Liang, T., et~al.
\newblock Mvimgnet: A large-scale dataset of multi-view images.
\newblock In \emph{Proceedings of the IEEE/CVF conference on computer vision
  and pattern recognition}, pp.\  9150--9161, 2023{\natexlab{b}}.

\bibitem[Zhang et~al.(2022{\natexlab{a}})Zhang, Zhang, Pham, Niu, Qiao, Yoo,
  and Kweon]{zhang2022dual}
Zhang, C., Zhang, K., Pham, T.~X., Niu, A., Qiao, Z., Yoo, C.~D., and Kweon,
  I.~S.
\newblock Dual temperature helps contrastive learning without many negative
  samples: Towards understanding and simplifying moco.
\newblock In \emph{Proceedings of the IEEE/CVF Conference on Computer Vision
  and Pattern Recognition}, pp.\  14441--14450, 2022{\natexlab{a}}.

\bibitem[Zhang et~al.(2022{\natexlab{b}})Zhang, Zhang, Zhang, Pham, Yoo, and
  Kweon]{zhang2022how}
Zhang, C., Zhang, K., Zhang, C., Pham, T.~X., Yoo, C.~D., and Kweon, I.~S.
\newblock How does simsiam avoid collapse without negative samples? a unified
  understanding with self-supervised contrastive learning.
\newblock In \emph{International Conference on Learning Representations},
  2022{\natexlab{b}}.
\newblock URL \url{https://openreview.net/forum?id=bwq6O4Cwdl}.

\bibitem[Zhang et~al.(2023)Zhang, Rao, and Agrawala]{zhang2023adding}
Zhang, L., Rao, A., and Agrawala, M.
\newblock Adding conditional control to text-to-image diffusion models, 2023.

\bibitem[Zheng et~al.(2019)Zheng, Song, Chen, Hu, Cao, and
  Nie]{zheng2019virtually}
Zheng, N., Song, X., Chen, Z., Hu, L., Cao, D., and Nie, L.
\newblock Virtually trying on new clothing with arbitrary poses.
\newblock In \emph{Proceedings of the 27th ACM international conference on
  multimedia}, pp.\  266--274, 2019.

\end{thebibliography}
\bibliographystyle{conferences}

%%%%%%%%%%%%%%%%%%%%%%%%%%%%%%%%%%%%%%%%%%%%%%%%%%%%%%%%%%%%%%%%%%%%%%%%%%%%%%%
%%%%%%%%%%%%%%%%%%%%%%%%%%%%%%%%%%%%%%%%%%%%%%%%%%%%%%%%%%%%%%%%%%%%%%%%%%%%%%%
% APPENDIX
%%%%%%%%%%%%%%%%%%%%%%%%%%%%%%%%%%%%%%%%%%%%%%%%%%%%%%%%%%%%%%%%%%%%%%%%%%%%%%%
%%%%%%%%%%%%%%%%%%%%%%%%%%%%%%%%%%%%%%%%%%%%%%%%%%%%%%%%%%%%%%%%%%%%%%%%%%%%%%%
\newpage
\appendix
\onecolumn

% You can have as much text here as you want. The main body must be at most $8$ pages long.
% For the final version, one more page can be added.
% If you want, you can use an appendix like this one.  

% The $\mathtt{\backslash onecolumn}$ command above can be kept in place if you prefer a one-column appendix, or can be removed if you prefer a two-column appendix.  Apart from this possible change, the style (font size, spacing, margins, page numbering, etc.) should be kept the same as the main body.
%%%%%%%%%%%%%%%%%%%%%%%%%%%%%%%%%%%%%%%%%%%%%%%%%%%%%%%%%%%%%%%%%%%%%%%%%%%%%%%
%%%%%%%%%%%%%%%%%%%%%%%%%%%%%%%%%%%%%%%%%%%%%%%%%%%%%%%%%%%%%%%%%%%%%%%%%%%%%%%

\section{Appendix}
\label{appendix_sec}
\subsection{More details of experimental setups}
We use 50 steps DDIM \cite{song2020denoising} for inference which is the same as AnyDoor \cite{chen2024anydoor}. The details of our method's configuration are provided in Tab. \ref{tab:configuration}. For VAE, we used the VAE of Stable Diffusion \cite{rombach2022high}.
\begin{table}[!htbp]
    \caption{\textbf{Parameters and Configs}. We follow ViT \cite{dosovitskiy2010image} to name models Large (L). }
    \label{tab:configuration}
    \begin{center}
    \resizebox{1\hsize}{!}{
    \begin{tabular}{ccccc|ccccc|ccccc}
    \toprule
    Method & Layers & Dim. & Heads & Param. (M) & Method & Layers & Dim. & Heads & Param. (M) & Method & Layers & Dim. & Heads & Param. (M) \\
    \midrule
    DiT-L & 24 & 1024 & 16 & 458.0 & MDT-L & 24 & 1024 & 16 & 459.1 & E-MD3C & 24 & 1024 & 16 & 468.0 \\
    \bottomrule
    \end{tabular}}
    \end{center}
\end{table}

\subsection{Datasets Used for Training}
\label{sec:appendix_dataset}
We utilize publicly available datasets, comprising a mix of image and video datasets, similar to AnyDoor \cite{chen2024anydoor}. However, due to challenges in downloading and processing, some datasets could not be included (`\xmark' as denoted in the table). The datasets used for training our model are detailed in Tab. \ref{tab:appendix_dataset}.

\begin{table}[!htbp]
    \caption{\textbf{Training Dataset}. All datasets marked in `\cmark' are used for training our model within the column ``Ours Used''.}
    \label{tab:appendix_dataset}
    \begin{center}
    \resizebox{1\hsize}{!}{
    \begin{tabular}{ccccccc}
    \toprule
    Dataset & Type & \# Samples & Variation & Quality & AnyDoor Used & Ours Used \\
    \midrule
    YouTubeVOS \cite{xu2018youtube} & Video & 4,450 & \cmark & Low & \cmark & \cmark \\
    YouTubeVIS \cite{yang2019video} & Video & 2,883 & \cmark & Low & \cmark & \xmark \\
    UVO \cite{wang2021unidentified} & Video & 10,337 & \cmark & Low & \cmark & \xmark \\
    MOSE \cite{ding2023mose} & Video & 1,507 & \cmark & High & \cmark & \cmark \\
    VIPSeg \cite{miao2022large} & Video & 3,110 & \cmark & High & \cmark & \cmark \\
    BURST \cite{athar2023burst} & Video & 1,493 & \cmark & Low & \cmark & \xmark \\
    \midrule
    MVImgNet \cite{yu2023mvimgnet} & Multi-view Image & 104,261 & \cmark & High & \cmark & \cmark \\
    VITON-HD \cite{choi2021viton} & Multi-view Image & 11,647 & \cmark & High & \cmark & \cmark \\
    FashionTryon \cite{zheng2019virtually} & Multi-view Image & 21,197 & \cmark & High & \cmark & \cmark \\
    \midrule
    MSRA \cite{borji2015salient} & Single Image & 10,000 & \xmark & High & \cmark & \cmark \\
    DUT \cite{wang2017learning} & Single Image & 15,572 & \xmark & High & \cmark & \cmark \\
    Flickr \cite{cong2020dovenet} & Single Image & 4,833 & \xmark & High & \cmark & \cmark \\
    LVIS \cite{gupta2019lvis} & Single Image & 117,287 & \xmark & High & \cmark & \cmark \\
    SAM (subset) \cite{kirillov2023segment} & Single Image & 100,864 & \xmark & High & \cmark & \cmark \\
    \bottomrule
    \end{tabular}}
    \end{center}
\end{table}

\subsection{Self-Supervised Learning Models}
There are various SSL models have been explored to learn the representations without labels \cite{he2022masked, pham2021self, pham2023self, oquab2023dinov2, zhang2022dual, zhang2022how}. These models serve as a good extractor for various applications \cite{pham2022lad, chen2024anydoor}. DINOv2 \cite{oquab2023dinov2} demonstrated an excellent pre-trained model for various diffusion-based frameworks. We mainly use DINOv2, but the other options may be worth trying. With the potential of diffusion transformers for conditional learning, it is expected to have more discovery of its capability in various domains and applications such as speech processing \cite{jung2022deep, jung2020flexible, trungshort}, data augmentation \cite{lee2020learning}, VQA \cite{kim2020modality}, visual detection learning \cite{vu2019cascade}, super-resolution \cite{niu2023cdpmsr,niu2024acdmsr,niu2024learning}.

\subsection{More Quantitative Results}
\label{sec:appendix_objects}
\begin{figure}[!htbp]
  \centering
  \includegraphics[width=1\linewidth]{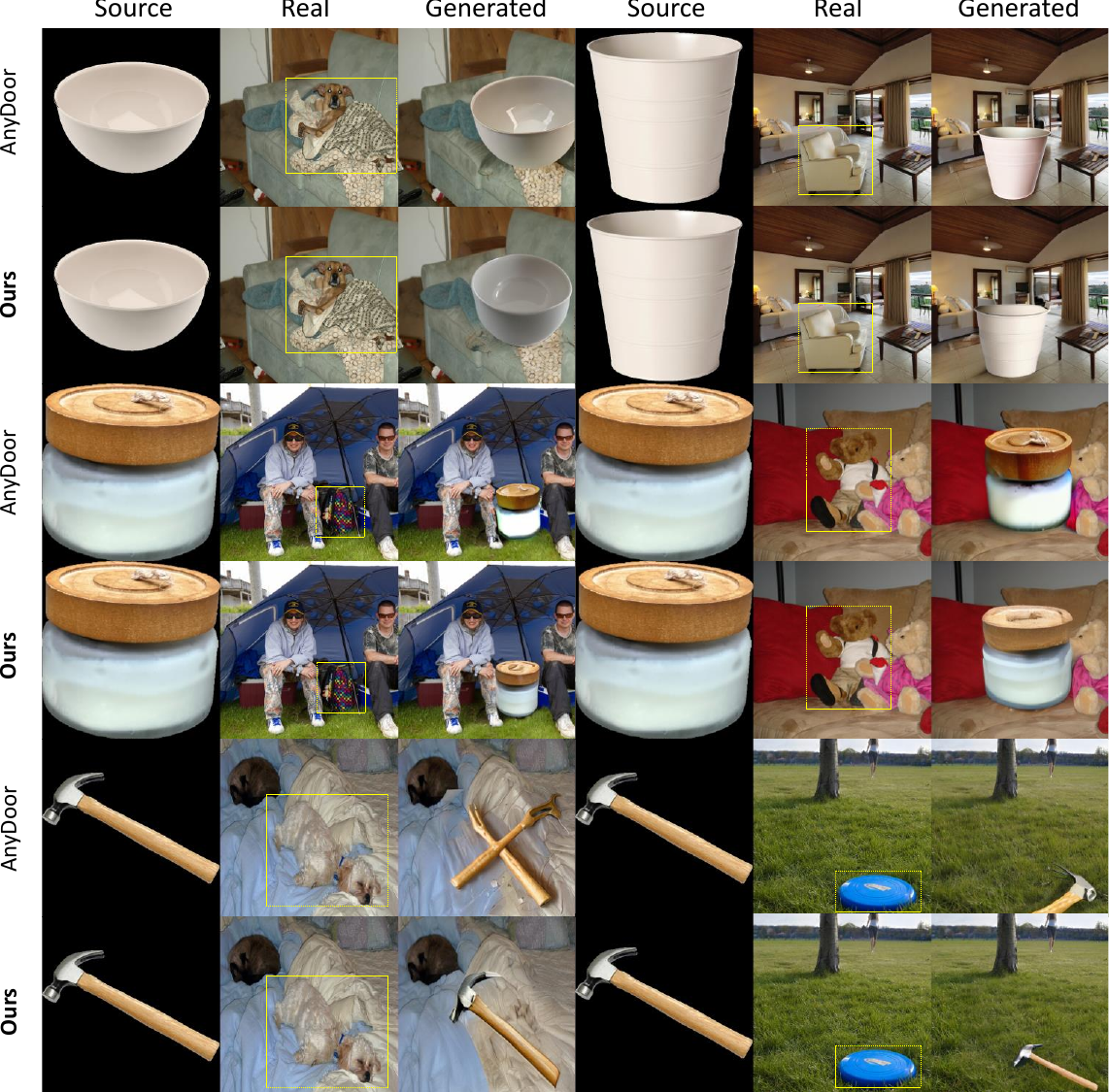}
  \caption{ \textbf{Object Composition.} Compared existing work AnyDoor and ours E-MD3C (1).
  }
  \label{fig:appendix_obj1}
\end{figure}

\begin{figure}[!htbp]
  \centering
  \includegraphics[width=1\linewidth]{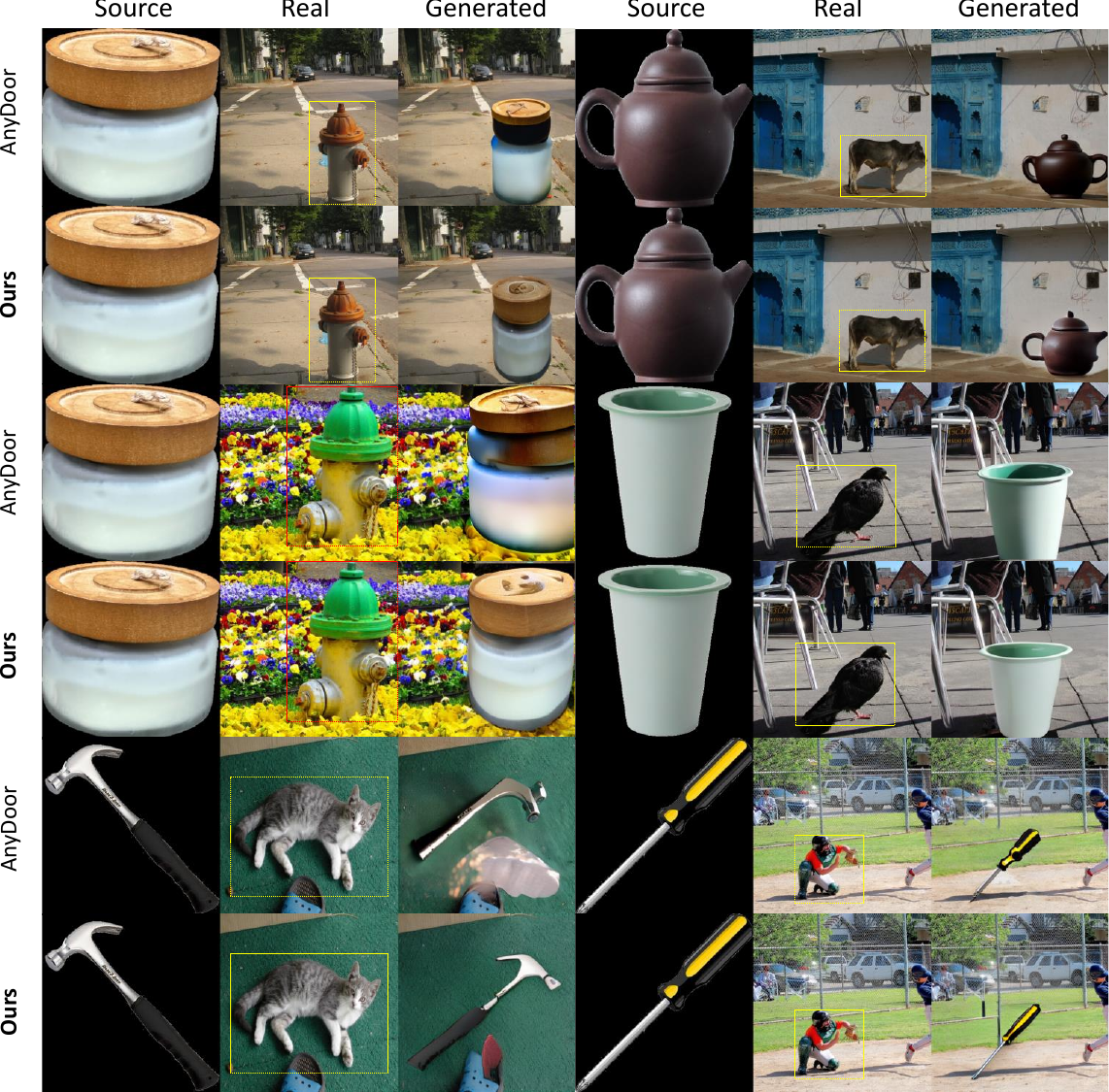}
  \caption{ \textbf{Object Composition.} Compared existing work AnyDoor and ours E-MD3C (2).
  }
  \label{fig:appendix_obj2}
\end{figure}

\begin{figure}[!htbp]
  \centering
  \includegraphics[width=1\linewidth]{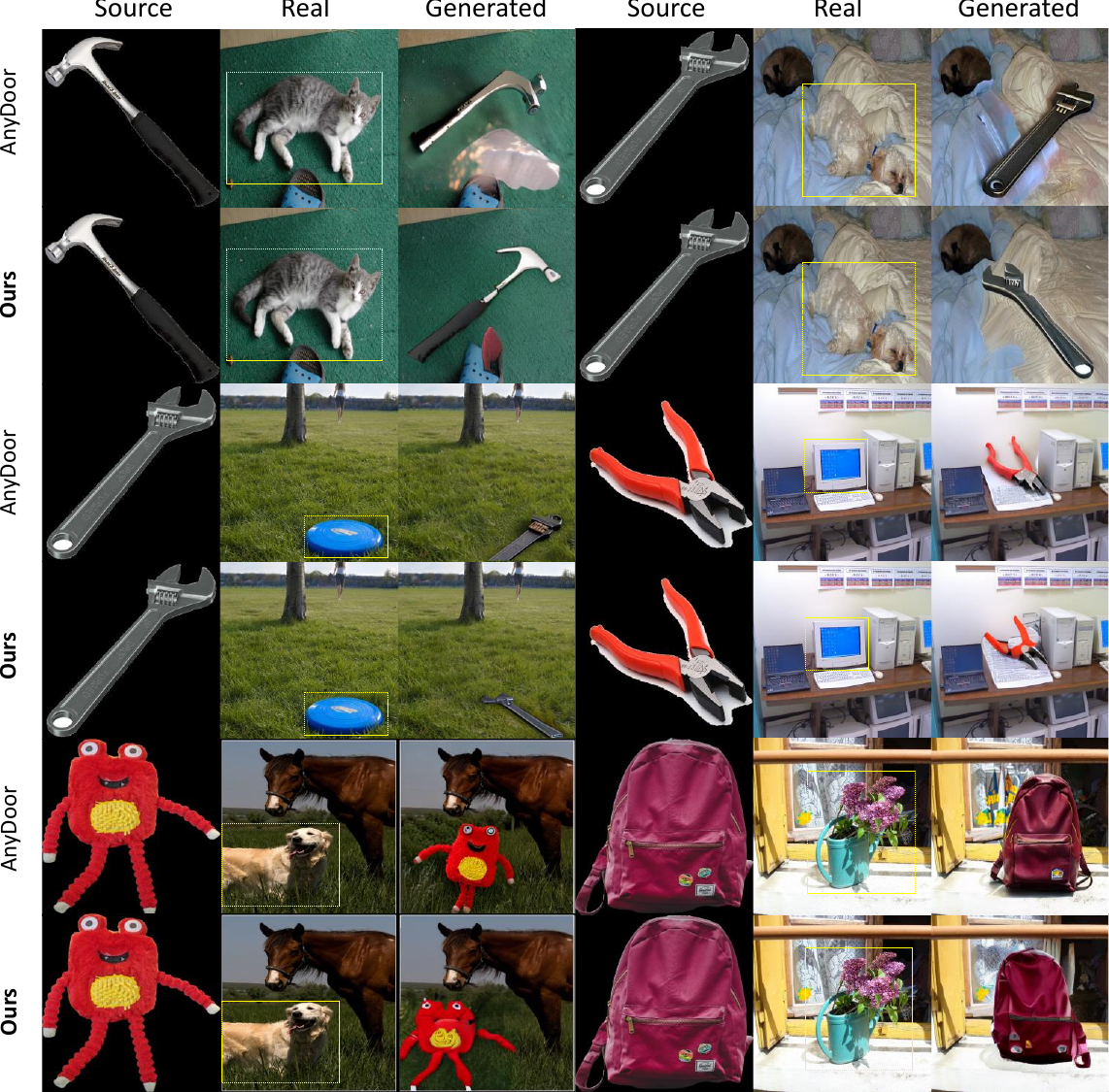}
  \caption{ \textbf{Object Composition.} Compared existing work AnyDoor and ours E-MD3C (3).
  }
  \label{fig:appendix_obj3}
\end{figure}

We demonstrate the effectiveness of our method in various object composition tasks, as illustrated in Fig. \ref{fig:appendix_obj1}, Fig. \ref{fig:appendix_obj2}, and Fig. \ref{fig:appendix_obj3}.

\end{document}